\documentclass[11pt]{article}

\usepackage[final]{acl}

\usepackage{times}
\usepackage{latexsym}
\usepackage{multirow}
\usepackage{verbatim}
\usepackage{amsmath}
\usepackage{cleveref}
\usepackage{enumitem}
\usepackage{kotex}
\usepackage{amssymb}
\usepackage{xspace}
\usepackage{booktabs}
\usepackage{multirow}
\usepackage[table]{xcolor}
\newcommand{\dd}[1]{\,\textsubscript{\textcolor{red}{$\downarrow$#1}}}
\newcommand{\uu}[1]{\,\textsubscript{\textcolor{teal}{$\uparrow$#1}}}
\usepackage{tcolorbox}
\usepackage{enumitem}
\usepackage{caption}
\usepackage{makecell}
\usepackage{multirow}
\usepackage{booktabs}
\usepackage{graphicx}
\usepackage{amssymb}
\usepackage{algorithm}
\usepackage{algpseudocode}
\usepackage{makecell}
\usepackage{placeins}

\usepackage[T1]{fontenc}

\usepackage[utf8]{inputenc}

\usepackage{microtype}

\usepackage{inconsolata}

\usepackage{graphicx}

\newcommand{\algname}{PGMem}

\definecolor{revisionblue}{RGB}{0,90,180}

\newif\ifshowrevisions
\showrevisionsfalse    

\ifshowrevisions
  \newcommand{\rev}[1]{{\color{revisionblue}#1}}
\else
  \newcommand{\rev}[1]{#1}
\fi

\algnewcommand{\LComment}[1]{%
  \State \(\triangleright\) \textit{#1}
}

\definecolor{oursrowlight}{RGB}{222,231,228}
\definecolor{oursrow}{RGB}{194,207,202}

\title{PGMem: Tightly Coupled Persona--Memory Graph for Lifelong Personalized Agents}

\author{
  Wonjun Choi\textsuperscript{1} \quad
  Yerim Kim\textsuperscript{1} \quad
  Yukyung Lee\textsuperscript{2,\ensuremath{\dagger}} \quad
  Susik Yoon\textsuperscript{1,\ensuremath{\dagger}} \\
  \textsuperscript{1}Korea University, Seoul, Korea \\
  \textsuperscript{2}Boston University, Boston, USA \\
  \texttt{\{migreeni, dpfla274, susik\}@korea.ac.kr},
  \texttt{ylee5@bu.edu}
}

\begin{document}
\maketitle
\begingroup
\renewcommand{\thefootnote}{}
\footnotetext{%
  \textsuperscript{\ensuremath{\dagger}}Corresponding authors.
}
\endgroup
\begin{abstract}
Long-term personalized dialogue agents must track user preferences as their personas evolve.
Existing memory systems organize past events well, but store personas as flat profiles detached from the events that justify them.
This loose coupling leads to the memory--persona validity gap and the persona-aware retrieval gap. We propose \algname{}, a heterogeneous persona-memory graph that connects event and persona nodes through typed provenance and evidence edges, keeping each persona signal traceable to the events that support or revise it.
At retrieval time, \algname{} expands from query-relevant seeds and ranks signals by evidential validity.
    Across three benchmarks with small language model backbones, \algname{} consistently outperforms summary-based, persona-aware, graph-structured, and agentic memory baselines, and improves performance as the context grows.  The source code of \algname{} is available at \url{https://github.com/wonjunchoi23/pgmem}.
\end{abstract}

\section{Introduction}
\label{sec:intro}

Lifelong AI agents are increasingly expected to provide personalized interactions as large language models (LLMs) continue to advance~\citep{lifelongagent26}. 
Building such agents requires a memory module that organizes historical interactions and tracks the evolution of user personas~\citep{memorysurvey25}. 
This dependence on explicit memory becomes even more pronounced for agents on small language models (SLMs)~\citep{slmsurvey25}, whose limited context windows and reasoning capabilities restrict their ability to recover such information directly from raw dialogue.

One primary line of relevant work has advanced by improving memory construction and retrieval for long-term dialogue. Early methods stored past conversations as summaries or compressed memories~\citep{memorybank24, rsum25, comedy25}. Subsequent work moved beyond flat compression by introducing hierarchical context management~\citep{memgpt23}, segment-level memory~\citep{secom25}, and agentic memory organization~\citep{amem25}.
More recent graph-based memory frameworks~\citep{theanine25, hypermem26, fracom25} make relations among these event-centric units explicit, capturing related memories beyond isolated top-$k$ similarity retrieval. 
While they employ structurally richer memories, the units they organize remain mere records of what was said and what happened, leaving the user's evolving persona outside the formal structure.

In parallel, personalization-oriented approaches introduce user profiles or portraits~\citep{msp22, memorybank24} or dedicated persona modules~\citep{dulemon22, ldagent25} to incorporate user-level representations into response generation. However, these methods keep the persona as a condensed user description or a separate persona bank, leaving it structurally detached from the event memories. Even when updated, such flat stores fail to track whether a persona signal still remains valid when later interactions conflict with it.

Taken together, existing approaches treat memory and persona as \textit{loosely coupled} components, enabling partial personalization but falling short of truly personalized lifelong interaction. Consequently, neither line of work explicitly models how events and persona signals ground one another.
This structural decoupling drives a common failure mode, attributed to two fundamental gaps.

\begin{figure}[t]
\includegraphics[width=\columnwidth]{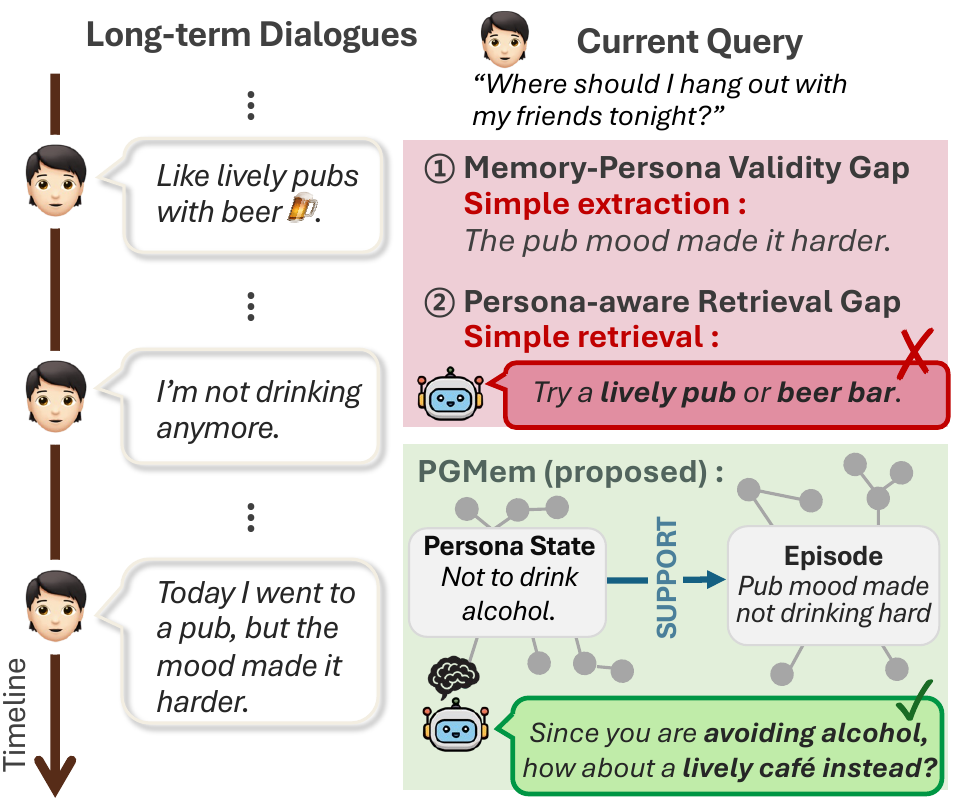}
\vspace{-0.2cm}
\caption{A motivating example of personalization failures with shifted user persona across sessions. }
\vspace{-0.2cm}
\label{motivate_fig}
\end{figure}

As illustrated in Figure~\ref{motivate_fig}, consider a scenario where a user first expresses a preference for lively pubs with beer, later states that they have stopped drinking, and subsequently reports that visiting a pub made this change harder. 
This final utterance carries a persona signal only implicitly. 
In this context, the difficult pub visit becomes informative only in light of the earlier shift, as the user is now trying to abstain. 
If stored merely as a surface-level episode, this evidence fails to validate the user's current non-drinking state. 

Specifically, the two gaps inherent in existing approaches are summarized as:
\begin{itemize}[leftmargin=*]
\setlength{\itemsep}{0pt}
\item The \emph{memory--persona validity gap} at memory construction time. 
Existing methods rarely store episodic memories and persona signals as coupled evidence, leaving them unable to represent why a persona signal emerged, how later events support it, or whether a newer signal revises an earlier preference.
\item The \emph{persona-aware retrieval gap} at query time. 
Existing methods rely heavily on surface-level query--memory similarity, omitting latent persona states from the retrieval step. Consequently, they surface highly similar historical episodes that answer the literal query but directly conflict with the user's updated persona, failing to capture implicit, weakly matched contexts that are essential for persona-consistent generation. 
\end{itemize}

To address these gaps, we propose \textbf{\algname{}}, a heterogeneous persona-memory graph for long-term personalized dialogue.
The core idea of \algname{} is to \textit{tightly couple memory and persona} by framing them as multiple layers of a single graph linked by explicit evidential relations.
Lifelong personalization thus shifts from mere profile accumulation to proactive evidence management: a persona signal guides response generation only when it is grounded in dialogue evidence and remains consistent with subsequent interactions.

\algname{} implements this idea as a heterogeneous graph with two coupled levels of nodes and typed edges.
The event-level contains Context nodes for raw utterance pairs and Episode nodes for episodic summaries.
The persona-level contains State nodes for transient persona signals and Trait nodes for stable dispositions.
Simultaneously, these nodes are interconnected by two distinct edge families.
Source edges encode provenance across abstraction levels, whereas Evidence edges encode evidential relations between memory and persona: support, contradiction, and temporal shift.
This structure fundamentally reshapes retrieval.
\algname{} starts from query-relevant seed nodes and expands along evidence edges, reaching persona signals that the query barely matches on the surface.
\algname{} then scores each retrieved signal by its evidential validity.
Outdated or contradicted signals are down-weighted, while currently valid persona evidence is prioritized for response generation.

Revisiting Figure~\ref{motivate_fig}, this tightly coupled framework resolves both gaps.
During memory construction, the difficult pub visit is linked to the newly updated non-drinking state as supporting evidence.
At retrieval time, expansion along evidence edges surfaces the validated non-drinking state and prevents the outdated pub preference from influencing the response.
Finally, the agent suggests an alcohol-free venue rather than a lively bar, enabling persona-consistent generation.

Our contributions are summarized as follows:
\begin{itemize}[leftmargin=*]
\setlength{\itemsep}{0pt}
\item To our knowledge, this is the first work to formulate lifelong personalized dialogue as a problem of evidence-grounded persona management, moving beyond merely accumulated user profiles and characterizing two structural gaps: the memory--persona validity gap and the persona-aware retrieval gap.

\item We propose \algname{}, a heterogeneous persona-memory graph that tightly couples event memory with persona signals. Typed edges ground persona signals in dialogue evidence and enable validity-aware retrieval.

\item We evaluate \algname{} on three long-term personalized dialogue benchmarks under SLM backbones. 
\algname{} outperforms memory- and persona-based baselines, and its advantage widens as dialogue context grows. Ablations confirm the contribution of persona nodes, evidence edges, and validity-aware retrieval.

\end{itemize}
\section{Related Work}

\paragraph{Memory construction and management.} Prior work on long-term dialogue memory represents dialogue history as summaries or compact memory states, focusing on how past interactions can be compressed and maintained over time~\citep{memorybank24,rsum25,comedy25}.
Subsequent work further shows that memory granularity matters, motivating segment-level or multi-granularity designs~\citep{secom25,rmm25}.
Another thread addresses memory maintenance through hierarchical management, offline consolidation, and agentic evolution~\citep{memgpt23,mem025,lightmem25,meminsight26,amem25}.
These methods advance memory organization yet leave persona signals outside the structure. \algname{} instead links them to their episodic evidence through typed provenance and evidential edges.

\paragraph{Graph-based memory.} Graph-based memory systems connect memories through temporal, causal, semantic, or event-centric relations and retrieve evidence through graph connectivity rather than independent similarity ranking~\citep{theanine25, associa25, sgmem25, magma26}. FraCom~\citep{fracom25} decomposes dialogue into proposition fragments for graph-based composition at retrieval time, while HyperMem~\citep{hypermem26} introduces a topic-episode-fact hypergraph to capture higher-order associations beyond pairwise edges.
Existing graph memories use edges mainly to expand retrieval over related events or content units. \algname{} instead uses evidence paths to assess which persona signals remain valid for generation.

\paragraph{Personalization.} Personalization-oriented systems explicitly maintain user-level representations for long-term dialogue. Some systems maintain user portraits or user--bot dynamics as part of memory representations~\citep{memorybank24,comedy25}, while others introduce dedicated persona extraction modules alongside event memory~\citep{ldagent25} or heterogeneous memory designs for personalized assistant settings~\citep{mempal26}.
Yet persona is typically stored as profile-like summaries or descriptor banks, detached from its episodic evidence.
\algname{} replaces flat persona profiles with individual signals whose validity is assessed against accumulating dialogue evidence.

\section{Problem Setting}
\label{sec:task_definition}

We consider long-term personalized dialogue, in which an assistant interacts with a single user over an extended sequence of utterances. Let $u_t^{\rho_t}$ denote the $t$-th utterance, where $u_t$ is the utterance content and $\rho_t \in \{\mathrm{user}, \mathrm{assistant}\}$ indicates the speaker role. The utterance-level dialogue history up to step $t$ is denoted as:

\begin{equation}
\mathcal{H}_{\leq t} = (u_1^{\rho_1}, u_2^{\rho_2}, \ldots, u_t^{\rho_t}).
\end{equation}

The system maintains an external memory state $\mathcal{M}^{(t)}$ that organizes user-relevant information derived from $\mathcal{H}_{\leq t}$, including a user persona state $\mathcal{P}^{(t)} \subseteq \mathcal{M}^{(t)}$. Given a user query $q_t$ at response time $t$, the system first retrieves relevant memories from the previous memory state, conditioned on the user's persona state:

\begin{equation}
\hat{\mathcal{M}}_{q_t} = \mathcal{R}(q_t, \mathcal{M}^{(t-1)} ; \mathcal{P}^{(t-1)}).
\end{equation}

The retrieved memory set $\hat{\mathcal{M}}_{q_t}$ then conditions response generation, again together with the persona state:
\begin{equation}
r_t = f_{\mathrm{LM}}(q_t, \hat{\mathcal{M}}_{q_t} ; \mathcal{P}^{(t-1)}).
\end{equation}

The memory $\mathcal{M}^{(t)}$ is updated incrementally as the dialogue unfolds, evolving $\mathcal{P}^{(t)}$ along with it:
\begin{equation}
\mathcal{M}^{(t)} = \mathcal{U}(\mathcal{M}^{(t-1)}, u_t^{\rho_t}),
\quad
\mathcal{M}^{(0)}=\emptyset.
\end{equation}

In this work, we instantiate $\mathcal{M}$ as a heterogeneous persona-memory graph (\cref{sec:graph_structure}), $\mathcal{U}$ as incremental graph construction (\cref{sec:construction}), and $\mathcal{R}$ as evidence-guided graph retrieval (\cref{sec:retrieval}).

\section{The Proposed Framework \algname{}}
\begin{figure*}[t!]
  \centering  
  \includegraphics[width=\textwidth]{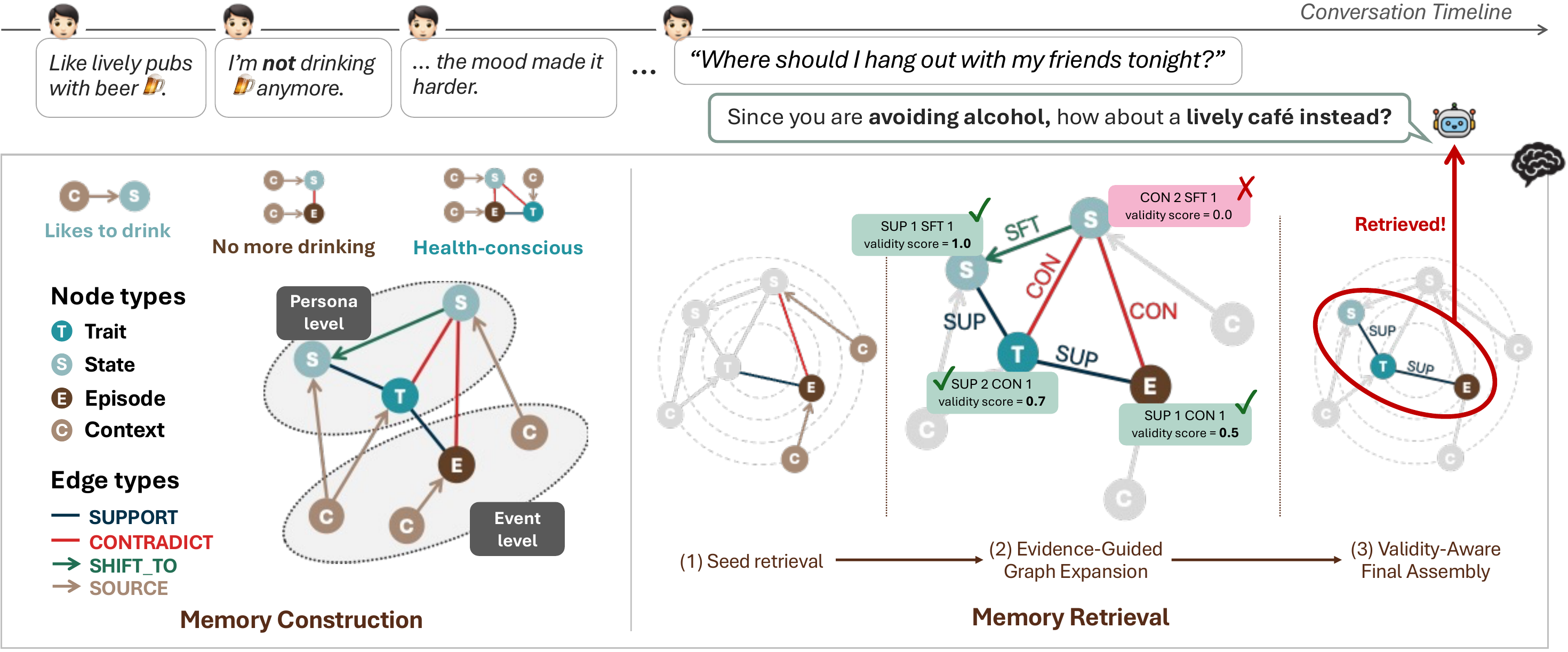} 

\caption{
  Overview of \algname{}. \emph{Memory Construction} builds a heterogeneous persona-memory graph from dialogue, with event-level and persona-level nodes connected by \textsc{source} edges and typed evidence edges.
  \emph{Memory Retrieval} answers a query in three stages: (1) seed retrieval selects query-aligned anchors, (2) evidence-guided graph expansion traverses typed edges to recover indirectly relevant evidence, and (3) validity-aware final assembly ranks the expanded pool by query relevance and evidential validity.
}
  \label{fig:gmem-framework}
\end{figure*}

As illustrated in Figure \ref{fig:gmem-framework}, \algname{} employs a heterogeneous persona-memory graph for long-term personalized dialogue that tightly couples event records with persona signals through explicit evidential relations.

\subsection{Heterogeneous Persona-Memory Graph}
\label{sec:graph_structure}
\algname{} structures dialogue history as a heterogeneous persona-memory graph ${\mathcal{G}=(\mathcal{V}, \mathcal{E})}$. Specifically, the node set $\mathcal{V}$ comprises two complementary families: \emph{event-level nodes} which capture episodic contexts, and \emph{persona-level nodes} which track the user's evolving state and traits. While these two families remain structurally distinct, explicit cross-type relations $\mathcal{E}$ keep persona signals traceable to the underlying source and evidence that validates or revises them.


\subsubsection{Heterogeneous Nodes}
\label{sec:heterogeneous_graph}
The node set $\mathcal{V}$ is organized along two complementary levels.
The \emph{event-level} ($\mathcal{V}^{c}$, $\mathcal{V}^{e}$) preserves what was said and what happened, while the \emph{persona-level} ($\mathcal{V}^{s}$, $\mathcal{V}^{t}$) maintains hypotheses about who the user is, mirroring the cognitive distinction between episodic memory~\citep{episodic72} and state--trait personality representations~\citep{statetotrait01,bigfive99}.

\paragraph{Event-level: Context and Episode.}
A \emph{Context} node $v^{c} \in \mathcal{V}^{c}$ stores the raw utterance at each dialogue turn. 
An \emph{Episode} node $v^{e} \in \mathcal{V}^{e}$ summarizes a chunk of consecutive dialogue turns into an event-level representation of the local conversation.

\paragraph{Persona-level: State and Trait.}
A \emph{State} node $v^{s} \in \mathcal{V}^{s}$ captures a short-term persona signal expressed or implied at a specific point in dialogue, such as a current preference, constraint, goal, stance, or situational condition. 
A \emph{Trait} node $v^{t} \in \mathcal{V}^{t}$ represents a more abstract persona hypothesis generalized from accumulated dialogues, capturing persistent regularities such as enduring preferences, values, habits, or behavioral tendencies.

\paragraph{Node attributes.}
A node $v \in \mathcal{V}$ is annotated with a keyword set $\mathcal{K}(v)$ and a domain-label set $\mathcal{D}(v)$, providing surface-level and abstract topical cues, respectively.
Categorical attributes \textsc{scope} and \textsc{recall\_priority} further qualify State nodes. 

\subsubsection{Provenance and Evidence Edges}
\label{sec:edges}
\paragraph{Source edges.}
$\mathcal{E}_{\mathrm{src}}$ encodes provenance across abstraction levels. Each source edge $v_i \twoheadrightarrow v_j$ denotes that the higher-level node $v_j$ is constructed from the lower-level node $v_i$.

\paragraph{Evidence edges.}
$\mathcal{E}_{\mathrm{evi}}$ encodes typed evidential relations among nodes. Each edge $(v_i,\,r,\,v_j)$ carries a label 
$r \in \mathcal{R}_{\mathrm{evi}} = \{\,\oplus,\; \ominus,\; \rightsquigarrow,\; \varnothing\,\}$, corresponding to \textsc{Support}, \textsc{Contradict}, 
\textsc{Shift\_To}, and \textsc{Irrelevant} relations, respectively. Each label simultaneously specifies the edge's evidential polarity and the path sign it contributes during graph expansion. For instance, consider a base State node $v^{s}_{0}$ representing \emph{``the user likes coffee''}:
\begin{itemize}[leftmargin=*]
\setlength{\itemsep}{0pt}
  \item $v_{i}\xrightarrow{\oplus}v_{j}$ \textsc{(Sup)}: compatible co-active evidence reinforcing $v_{j}$  (e.g., a Trait node \emph{``prefers strong espresso''} 
        $\xrightarrow{\oplus} v^{s}_{0}$).
  \item $v_{i}\xrightarrow{\ominus}v_{j}$ \textsc{(Con)}: conflicting co-active evidence marking $v_{j}$ as challenged (e.g., a co-active State node \emph{``trying to cut caffeine''} $\xrightarrow{\ominus} v^{s}_{0}$).
  \item $v_{i}\rightsquigarrow v_{j}$ \textsc{(Sft)}:  a temporally ordered, directional update with $v_{j}\succ_{t}v_{i}$, in which the newer signal $v_{j}$ supersedes $v_{i}$  (e.g., $v^{s}_{0}\rightsquigarrow$ \emph{``switched to tea''}).
  \item $v_{i} \overset{\emptyset}{\nleftrightarrow} v_{j}$ \textsc{(Irr)}: a judged but unrelated pair; carries no retrieval evidence but prevents repeated relation judgments during construction.
\end{itemize}

\subsection{Memory Graph Construction}
\label{sec:construction}

\algname{} instantiates the update operator $\mathcal{U}$ introduced in Section~\ref{sec:task_definition}. 
Given a new utterance $u_t$, the graph is updated by adding newly derived nodes and linking them to existing memory through provenance and evidence relations.

\paragraph{Node construction.}
Node construction follows the temporal granularity and abstraction level of each node type. 
Each dialogue turn yields a Context node for the raw user--assistant exchange. 
When the utterance carries a persona-relevant signal, a State node is also extracted.
At chunk boundaries, recent Context nodes are summarized into an Episode node that captures the local conversational event. 
Every $B$ chunks, accumulated Context, State, and Episode nodes are used to extract Trait nodes, which represent persistent persona hypotheses supported by repeated or stable evidence.
During extraction, each node is additionally annotated with retrieval attributes---a keyword set $\mathcal{K}(v)$, a domain-label set $\mathcal{D}(v)$, and, for State nodes, categorical attributes \textsc{scope} and \textsc{recall\_priority}---that later guide memory retrieval.

\paragraph{Evidence edge extraction.}
In the local update stage, newly created nodes are paired with recent and previously stored candidates, and each pair is assigned a relation label from $\mathcal{R}_{\mathrm{evi}}$.
The offline update stage extends this coverage by detecting long-range relations between distant, previously unconnected nodes.
Together, these locally and globally established edges form evidential paths that allow the retrieval stage to trace multi-hop evidential patterns across sparsely connected regions of the graph. 
See Algorithm 1 in Appendix~\ref{app:algorithm} for details.
\rev{The classification prompt applies a fixed decision priority that resolves uncertain pairs to \textsc{IRR}, so ambiguity contributes no retrieval evidence (Figure~\ref{fig:trait_state_relation_prompt}). 
Appendix~\ref{app:edge-cases} reports the resulting edge statistics, a human validation of the assigned labels, and representative labeling cases.}

\subsection{Memory Retrieval}
\label{sec:retrieval}

Given a user query $q$, \algname{} retrieves a compact evidence set from the memory graph.
Seed retrieval scores all candidate nodes across all four node types against $q$ by combining semantic and lexical similarity, selecting budgeted entry points per type.
\algname{} then expands this pool along typed evidence edges to recover supporting, conflicting, and shift-related evidence beyond direct query--memory relevance.
Final assembly ranks the expanded memory pool by query relevance and evidential validity, returning a validity-ranked persona-memory set.

\subsubsection{Stage 1: Seed Retrieval}
\label{sec:seed_retrieval}

Seed retrieval selects a pool of query-aligned anchors from the full node set, providing a controlled starting point for subsequent evidence-guided expansion.
Given a query $q$, \algname{} extracts a normalized keyword set $\mathcal{K}(q)$ at retrieval time.
The lexical overlap between the query and a node $v$ is
\begin{equation}
\mathrm{ov}(q,v) =
\frac{|\mathcal{K}(q) \cap (\mathcal{K}(v)\cup\mathcal{D}(v))|} {\max(|\mathcal{K}(q)|,1)},
\end{equation}
where both query keywords and node labels are normalized to single-word forms, making this a strict word-level match.

Each candidate node is scored by combining embedding cosine similarity $\mathrm{sem}(q,v)$ and lexical overlap $\mathrm{ov}(q,v)$:
\begin{equation}
\phi(q,v)
= w_{\mathrm{sem}}\,\cdot \mathrm{sem}(q,v) + w_{\mathrm{ov}}\,\cdot \mathrm{ov}(q,v).
\end{equation}
The weights $w_{\mathrm{sem}}$ and $w_{\mathrm{ov}}$ depend on $\textsc{scope}(v)\in\{\textsc{Broad},\textsc{Narrow}\}$; \textsc{Broad} nodes cover broader topical scope, so their lexical overlap is down-weighted to avoid spurious matches. 
Using $\phi(q,v)$, \algname{} selects the top-$k_{\mathrm{seed}}$ nodes from each of the four node types as seeds.

As a complementary signal, \algname{} maintains an Active Persona Set (APS): the top-$k_{\mathrm{aps}}$ State nodes marked with high \textsc{recall\_priority}, ranked by $\phi(q,v)$, which helps retain salient persona signals in the pool despite low query relevance.
The initial retrieval pool $\mathcal{W}_0$ is formed by combining the top-$k_{\mathrm{seed}}$ seeds per node type with the $k_{\mathrm{aps}}$ APS nodes.

\subsubsection{Stage 2: Evidence-Guided Expansion}
\label{sec:graph_expansion}

Direct query relevance alone fails to surface evidence that is structurally linked to retrieved persona signals but lacks topical overlap with the query.
\algname{} addresses this by tracing evidence edges outward from each seed, accumulating evidence signs along each path so that the expanded pool reflects not only query relevance but also the validity structure surrounding each retrieved signal.

\paragraph{Signed evidence traversal.}

Context seeds anchor query relevance but do not participate in expansion directly. They are replaced in the pool by their Source-linked Episode, State, and Trait nodes, completing the expansion origin.

Each evidence edge $(v_i,\,r,\,v_j)$ is mapped to a path sign 
$\sigma \in \{s^+,\,s^-\}$ before traversal:
$\oplus$\,(\textsc{sup}) and forward 
$\rightsquigarrow$\,(\textsc{sft}) edges carry~$s^+$;\;
$\ominus$\,(\textsc{con}) and backward 
$\rightsquigarrow$\,(\textsc{sft}) edges carry~$s^-$;\;
$\varnothing$\,(\textsc{irr}) edges are excluded.
Signs compose multiplicatively along each path:
\begin{equation}
  s^+\circ s^+=s^+, \qquad s^+\circ s^-=s^-.
\end{equation}
An $s^+$ edge adds compatible evidence and allows continued traversal, whereas an $s^-$ edge terminates the path, and in either case traversal stops once a path reaches the hop cap $H$.

\paragraph{Expanded pool.}
The expanded pool is \begin{equation}
\mathcal{W}_1 = \mathcal{W}_0 \cup \mathcal{V}^{\mathrm{exp}},
\end{equation}
where $\mathcal{V}^{\mathrm{exp}}$ denotes nodes discovered during evidence traversal. After deduplication, $\mathcal{W}_1$ is passed to final assembly.

\subsubsection{Stage 3: Validity-Aware Final Assembly}
\label{sec:final_assembly}
Given $\mathcal{W}_1$, \algname{} assembles a compact final evidence set by combining query relevance with evidential validity.
Superseded $\rightsquigarrow$\,(\textsc{sft}) sources whose targets also appear in~$\mathcal{W}_1$ are first removed.
For each remaining node $v \in \mathcal{W}_1$, let $\mathcal{S}(v)$ denote the set of pool nodes from which a path with cumulative sign~$s^+$ reaches~$v$, and $\mathcal{C}(v)$ the set from which a path with cumulative sign~$s^-$ reaches~$v$.
The validity score with a smoothing constant $\alpha$ is
\begin{equation}
\label{eq:validity-score}
\mathrm{val}(v) 
= \frac{|\mathcal{S}(v)|+\alpha}
       {|\mathcal{S}(v)|+|\mathcal{C}(v)|+2\alpha}.
\end{equation} 
The final ranking score is

\begin{equation}
\label{eq:final-score}
\psi(q,v)=\lambda\mathrm{val}(v)+(1-\lambda)\phi(q,v),
\end{equation}
where $\lambda$ is set to 0.5 by default.
\rev{The score uses only path counts, so a single mislabeled edge shifts $|S(v)|$ or $|C(v)|$ by one rather than flipping a decision (Appendix~\ref{app:edge-cases}).}

\algname{} selects the top-ranked nodes under a fixed budget.
The resulting evidence set---seeded by query relevance, expanded along evidential structure, and filtered by validity---is serialized by type and status into the generation prompt.

See Algorithm~\ref{alg:retrieval} in Appendix~\ref{app:algorithm} for details.

\section{Experiments}
\label{sec:experiments}
\subsection{Experimental Setup}

\paragraph{Benchmarks.}
We evaluate \algname{} on three long-term personalized dialogue benchmarks.
The \textit{opposed} subset of ImplexConv~\citep{implexconv25} is our primary stress test for implicit persona-aware retrieval, which requires reasoning over persona signals semantically distant from the query. 
PrefEval~\citep{prefeval25} targets a multi-persona setting; sessions from distinct personas are interleaved, testing whether the agent can correctly retrieve persona-specific knowledge at query time.
PersonaMem~\citep{personamem25} evaluates dynamic user profiling across temporally ordered sessions, including preference evolution and personalized recommendation.
\rev{The three benchmarks stress different axes of personalization. 
Together, they} evaluate whether memory mechanisms can capture temporally valid, user-specific, and persona-critical evidence beyond surface semantic relevance.

\paragraph{Baselines.}
We compare our method with representative long-term dialogue memory baselines. 
A Full-History baseline feeds all prior dialogue turns directly into the language model without external memory.
The remaining baselines are 
MemoryBank~\citep{memorybank24}, a summary-based long-term memory framework; 
LD-Agent~\citep{ldagent25}, a persona-aware personalized dialogue agent; 
THEANINE~\citep{theanine25}, a temporal graph-structured memory model; 
and A-MEM~\citep{amem25}, an agentic note-based evolving memory system. 
These baselines cover full-context, summary-based, persona-aware, graph-structured, and agentic memory paradigms.
\rev{We additionally evaluate SeCom~\citep{secom25}, a segment-level memory with compression-based denoising, and H$^2$Memory~\citep{mempal26}, a hierarchical heterogeneous memory framework, in Appendix~\ref{app:additional-baselines}.}

\paragraph{Implementation Details.}
We evaluate all methods using Qwen3-1.7B~\citep{qwen3_25} and Gemma-3-1B~\citep{gemma3_25} as backbone SLMs. We use all-MiniLM-L6-v2\footnote{\url{https://huggingface.co/sentence-transformers/all-MiniLM-L6-v2}} as the sentence embedding model for embedding-based memory operations and evaluation. All baselines and \algname{} are run under the same backbone and embedding settings.

\paragraph{Evaluation metrics.}
Evaluation follows the protocol suited to each benchmark's answer format. 
ImplexConv responses are evaluated with a four-dimensional binary checklist adapted from CheckEval~\citep{checkeval25}, a checklist-based LLM-as-a-judge method.
We further report human validation of the adapted ImplexConv checklist in Appendix~\ref{app:human_validation}.
PrefEval uses the LLM-as-a-judge framework provided by the original benchmark~\citep{prefeval25}. 
Both ImplexConv and PrefEval are thus scored on a 0-4 (5-point) scale, obtained by summing four binary dimension judgments.
PersonaMem is formulated as four-way multiple choice and evaluated by exact-match accuracy. We use gpt-4o-mini~\citep{gpt23} as the judge model at temperature $0$.

\begin{table}
\centering
\footnotesize
\renewcommand{\arraystretch}{1.3}
\setlength{\tabcolsep}{3pt}
\begin{tabular}{llcccc}
\toprule
\multirow{2}{*}{} 
& \multirow{2}{*}{\textbf{Method}} 
& \multirow{2}{*}{\textbf{ImplexConv}} 
& \multirow{2}{*}{\textbf{PrefEval}} 
& \multicolumn{2}{c}{\textbf{PersonaMem}} \\ 
\cline{5-6}
& & & & 32k & 128k \\ 
\hline
\multirow{6}{*}{\rotatebox{90}{\textbf{Qwen3-1.7B}}} 
& Full history & 0.99 & 1.32 & 40.58 & 34.14 \\
& MemoryBank   & \underline{1.03} & 1.21 & 41.26 & 36.56 \\
& LD-Agent     & 0.92 & 1.14 & 40.58 & 36.23 \\
& A-MEM        & 0.94 & \underline{1.39} 
               & \underline{45.33} & \underline{43.23} \\
& THEANINE     & 0.98 & 1.25 & 40.41 & 35.42 \\
\cline{2-6}
& \cellcolor{oursrowlight}\textbf{\algname{}}  
& \cellcolor{oursrowlight}\textbf{1.33} 
& \cellcolor{oursrowlight}\textbf{2.21} 
& \cellcolor{oursrowlight}\textbf{45.50} 
& \cellcolor{oursrowlight}\textbf{46.35} \\ 
\hline
\multirow{6}{*}{\rotatebox{90}{\textbf{Gemma-3-1B}}}
& Full history & 0.86 & 1.44 & 24.62 & 28.31 \\
& MemoryBank   & \underline{1.09} & \underline{1.79} 
               & 24.96 & 28.60 \\
& LD-Agent     & 0.89 & 1.61 
               & \underline{27.50} & \underline{29.85} \\
& A-MEM        & 0.57 & 1.60 & 24.11 & 29.48 \\
& THEANINE     & 0.98 & 1.55 & 26.99 & 29.59 \\
\cline{2-6}
& \cellcolor{oursrowlight}\textbf{\algname{}}  
& \cellcolor{oursrowlight}\textbf{1.41} 
& \cellcolor{oursrowlight}\textbf{2.03} 
& \cellcolor{oursrowlight}\textbf{30.39} 
& \cellcolor{oursrowlight}\textbf{34.18} \\ 
\bottomrule
\end{tabular}

\caption{
\label{tab:main-results}
Comparison results on long-term personalized dialogue benchmarks.
ImplexConv and PrefEval report LLM-as-a-judge scores for personalized
response quality on a 5-point scale (0--4), while PersonaMem reports
accuracy (\%) on user persona QAs.
\textbf{Bold} and \underline{underline} denote the best and second-best
results, respectively.
}
\end{table}

\begin{figure}[t]
\centering
\includegraphics[width=\columnwidth]{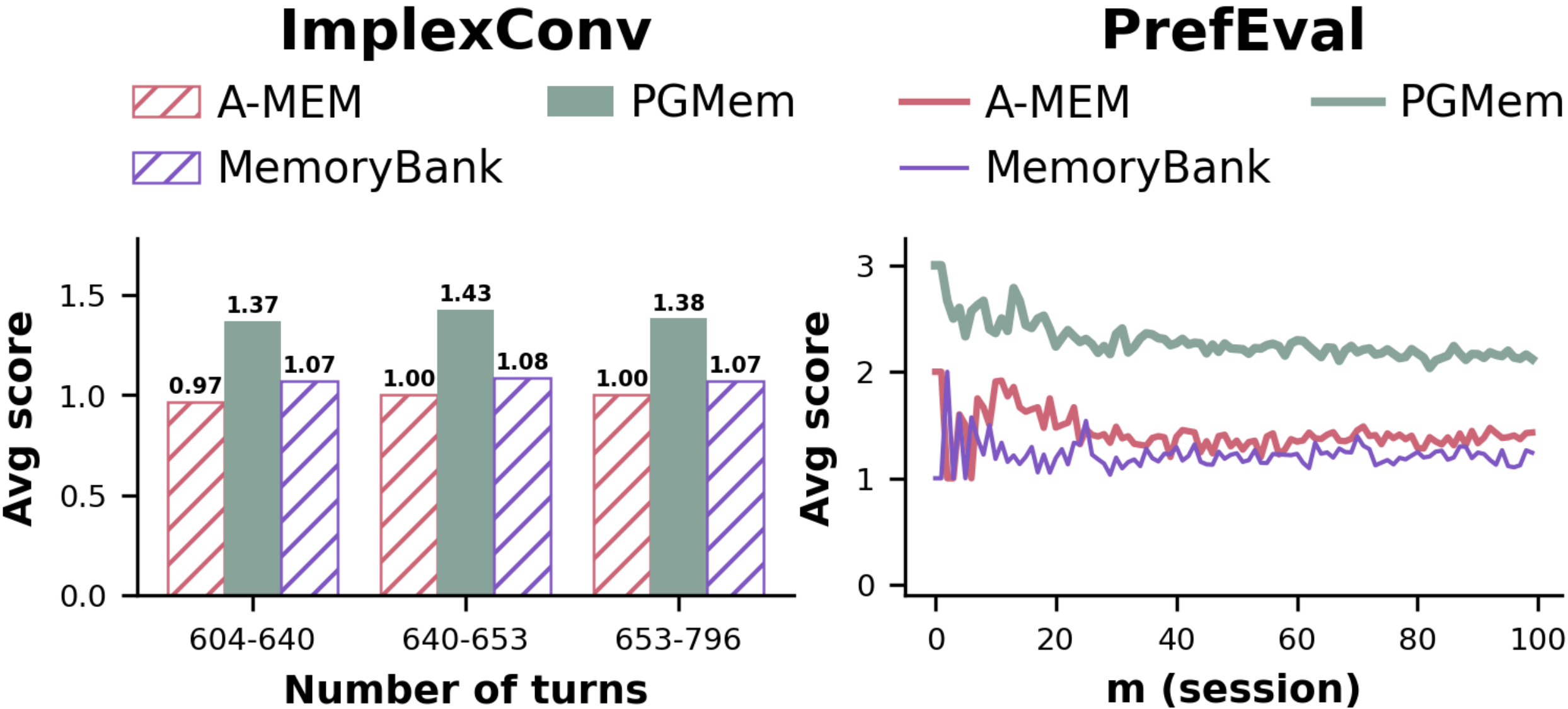}
\caption{Long-term consistency evaluation results. On ImplexConv (left), avg score is grouped by the number of accumulated dialogue turns. On PrefEval (right), avg score is tracked as the number of interleaved persona sessions $m$ grows.} 
\label{fig:longterm}
\end{figure}

\subsection{Overall Performance}

Table~\ref{tab:main-results} shows the results across three benchmarks under two backbone SLMs. \algname{} achieves the strongest personalization across all long-term dialogue settings.

\paragraph{Persona-aware retrieval.}
On ImplexConv, \algname{} shows the highest overall score, improving over the best baseline by up to 29\%, with particularly high scores on the persona-adaptation dimensions.
On PrefEval, \algname{} shows the highest score under both backbone SLMs (2.21 on Qwen3-1.7B and 2.03 on Gemma-3-1B), up to 59\% above the strongest baselines.

Baselines retrieve by query similarity alone, so they miss persona signals that the query depends on but does not topically overlap with. \algname{} recovers these signals through evidence-edge expansion, which its persona-adaptation scores directly reflect. The judge dimensions and per-dimension scores for both benchmarks are summarized in Appendices~\ref{app:evaluation} and~\ref{app:llm_judge_results}.

\paragraph{Long-term validity and consistency.}
In long-term dialogue the user's persona evolves over time, and the memory has to follow those shifts instead of letting stale signals pile up.
The long-horizon results from Figure~\ref{fig:longterm} indicate that \algname{} does this most consistently among the compared methods.
Across accumulating turns on ImplexConv and across interleaved persona sessions on PrefEval, \algname{} sustains a large and steady margin over the baselines.
On PersonaMem, \algname{}'s accuracy improves as the context window grows from 32k to 128k (45.5$\rightarrow$46.4 on Qwen3-1.7B, 30.4$\rightarrow$34.2 on Gemma-3-1B). 
On Qwen3-1.7B it is the only method to improve at all, while every baseline degrades. 
\algname{} achieves this with typed edges that tie each persona signal to its grounding events and a validity score that down-weights signals later interactions have superseded.

\begin{table}[t]
\centering
\small
\setlength{\tabcolsep}{4pt}
\renewcommand{\arraystretch}{1.3}
\resizebox{\columnwidth}{!}{%
\begin{tabular}{lcccc}
\toprule
\multirow{2}{*}{\textbf{Model}} & \multirow{2}{*}{\textbf{ImplexConv}} & \multirow{2}{*}{\textbf{PrefEval}} & \multicolumn{2}{c}{\textbf{PersonaMem}} \\
\cmidrule(lr){4-5}
 & & & \textbf{32k} & \textbf{128k} \\
\midrule
\textbf{\algname{}} & 1.33 & 2.21 & 45.50 & 46.35 \\
\midrule
\multicolumn{5}{l}{\textit{Component}} \\
\;w/o Persona-level & 1.21\dd{0.12} & 2.02\dd{0.19} & 41.09\dd{4.41} & 38.80\dd{7.55} \\
\;w/o Event-level   & 1.29\dd{0.04} & 2.24\uu{0.03} & 43.80\dd{1.70} & 42.28\dd{4.07} \\
\;w/o Graph         & 1.27\dd{0.05} & 2.21\dd{0.01} & 44.99\dd{0.51} & 43.53\dd{2.82} \\
\midrule
\multicolumn{5}{l}{\textit{Edge type}} \\
\;w/o $\oplus$\,(\textsc{sup})  & 1.33\dd{0.00} & 2.21\dd{0.00} & 45.84\uu{0.34} & 43.53\dd{2.82} \\
\;w/o $\ominus$\,(\textsc{con})  & 1.29\dd{0.04} & 2.20\dd{0.01} & 44.99\dd{0.51} & 44.37\dd{1.98} \\
\;w/o $\rightsquigarrow$\,(\textsc{sft})  & 1.31\dd{0.02} & 2.17\dd{0.04} & 45.33\dd{0.17} & 44.44\dd{1.91} \\
\bottomrule
\end{tabular}}
\caption{\rev{Ablation study of \algname{} on model components (top) and evidence edge types (bottom). Subscripts in red and blue indicate performance degradation and improvement relative to the full \algname{}, respectively.}}
\label{tab:ablation}
\end{table}

\subsection{Ablation Study}
\rev{We analyze the individual contributions of \algname{}'s components at two levels, covering its three main components (persona-level nodes, event-level nodes, and graph expansion) and its evidence relation types ($\oplus$\,(\textsc{sup}), $\ominus$\,(\textsc{con}), $\rightsquigarrow$\,(\textsc{sft})), under the same Qwen3-1.7B backbone.}
Table~\ref{tab:ablation} (top) shows that persona nodes are the most critical component: their removal yields the largest drop on every benchmark, confirming that persona-level representation drives \algname{}'s retrieval.
Event nodes contribute more modestly, with their removal even yielding a slight gain on PrefEval, indicating that Episode/Context nodes can occasionally introduce mild noise.
Removing graph expansion causes a degradation that widens sharply as dialogue history grows, confirming that the graph structure is what lets \algname{} surface valid persona evidence under long-term context.
\rev{Table~\ref{tab:ablation} (bottom) further ablates each evidence relation individually, where removing $\ominus$\,(\textsc{con}) or $\rightsquigarrow$\,(\textsc{sft}) consistently degrades performance on the benchmarks that test persona conflict and evolution, with the largest drops on PersonaMem-128k ($-$1.98 and $-$1.91, respectively), while $\oplus$\,(\textsc{sup}) matters primarily under the longest horizon.}

\subsection{In-Depth Analysis of \algname{}}

\begin{figure}[t]
\centering
\includegraphics[width=\columnwidth]{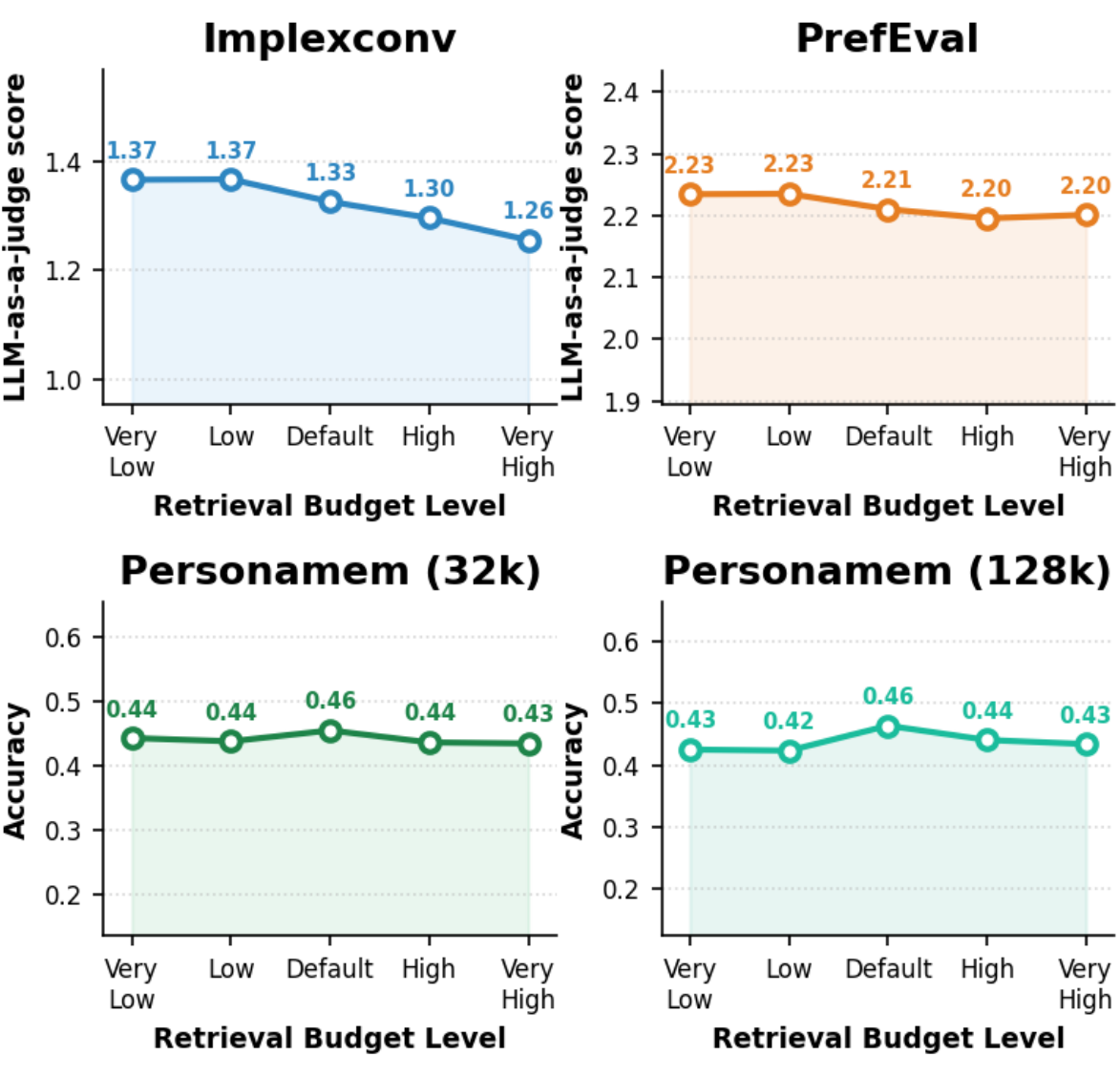}
\caption{Sensitivity to the retrieval budget, scaling the seed counts and final assembly sizes together around the default.}    
\label{fig:sensitivity}
\end{figure}

\paragraph{Sensitivity Analysis.}
We analyze \algname{}'s sensitivity to its retrieval budget \rev{and to the ranking weight $\lambda$ in Eq.~\eqref{eq:final-score}}. 
\rev{The budget} jointly controls the seed retrieval set before graph expansion and the final memory set passed to the generator.
The default budget is set to match \algname{}'s final memory size with the other baselines (full configuration in Appendix~\ref{app:hyperparameters}).
We scale every budget parameter together by a common factor across five levels from $0.3\times$ to $1.7\times$.
As shown in Figure~\ref{fig:sensitivity}, \algname{} remains stable across the entire range.
ImplexConv and PrefEval peak at the lower budget levels and decline mildly as the budget grows, while PersonaMem peaks at the default budget on both context lengths, making the default a reasonable compromise across all three benchmarks.
This matches our design rationale, in which graph expansion already recovers persona-relevant evidence beyond the seed set, so that a compact seed budget suffices and additional memories only introduce noise rather than useful evidence.

\begin{table}[t]
\centering
\small
\setlength{\tabcolsep}{4pt}
\renewcommand{\arraystretch}{1.3}
\begin{tabular}{lcccc}
\toprule
\multirow{2}{*}{$\boldsymbol{\lambda}$} & \multirow{2}{*}{\textbf{ImplexConv}} & \multirow{2}{*}{\textbf{PrefEval}} & \multicolumn{2}{c}{\textbf{PersonaMem}} \\
\cmidrule(lr){4-5}
 & & & \textbf{32k} & \textbf{128k} \\
\midrule
0.00 & 1.20 & 2.14 & 47.03 & 42.57 \\
0.25 & 1.28 & 2.22 & 46.35 & 43.34 \\
\textbf{0.50 (default)} & 1.33 & 2.21 & 45.50 & 46.35 \\
0.75 & 1.31 & 2.22 & 44.99 & 42.90 \\
1.00 & 1.34 & 2.23 & 43.97 & 43.12 \\
\bottomrule
\end{tabular}
\caption{\rev{Sensitivity to the ranking weight $\lambda$ in Eq.~\eqref{eq:final-score}, which balances evidential validity and query relevance.}}
\label{tab:lambda}
\end{table}

\rev{For $\lambda$, we sweep from 0 to 1 in steps of 0.25, with 0.5 as the default, and report results in Table~\ref{tab:lambda}. 
Removing validity from the ranking ($\lambda = 0$) is weakest on ImplexConv and PrefEval, so evidential validity contributes beyond query relevance. 
PersonaMem-32k favors query relevance and PersonaMem-128k peaks at the default, making $\lambda = 0.5$ a stable choice across all benchmarks. 
The weight is also an interpretable control over how strongly retrieval reflects the evolving persona.}

\paragraph{Efficiency Analysis.}
\label{sec:scalability}

\begin{table}[t]
\centering
\small
\setlength{\tabcolsep}{8pt}
\renewcommand{\arraystretch}{1.3}
\begin{tabular}{lccc}
\toprule
\multirow{2}{*}{\textbf{Method}} 
& \multicolumn{2}{c}{\textbf{Construction} ($10^3$ tok.)} 
& \textbf{QA} (tok.) \\
\cmidrule(lr){2-3} \cmidrule(lr){4-4}
 & \textbf{Input} & \textbf{Output} & \textbf{Input} \\
\midrule
MemoryBank   & 342.3   & 37.4  & 2,875  \\
LD-Agent     & 443.7   & 14.0  & 1,371  \\
A-MEM        & 2,265.7 & 633.4 & 5,124  \\
THEANINE     & 3,564.7 & 118.6 & 1,461  \\
Full History & --      & --    & 27,391 \\
\rowcolor{oursrowlight}
\textbf{\algname{}}  & 2,254.0 & 271.0 & 1,847  \\
\bottomrule
\end{tabular}
\caption{Token consumption on ImplexConv. Construction is the total tokens to build memory over a full session. QA is the average per query at inference. Dashes denote no construction stage (Full History).}
\label{tab:scaling}
\end{table}

Token consumption largely determines how a memory mechanism scales as dialogues grow longer. 
For ImplexConv, we measure token usage at the memory-construction stage, accumulated over a full session, and at the QA stage, incurred per query during inference. 
As shown in Table~\ref{tab:scaling}, at construction time, \algname{} builds a typed evidence graph yet incurs a lower total token cost than the other structure-building baselines, A-MEM and THEANINE.
At query time, \algname{} remains lightweight, staying well below the agentic and full-history baselines and only modestly above the cheapest ones. 
Summary-based methods are cheaper to construct but spend more per query.
\rev{Token count alone does not capture the operational cost of graph maintenance. In Appendix~\ref{app:latency}, we report LLM calls and wall-clock latency for memory construction, together with the number of LLM calls per query, for \algname{} and the two structure-building baselines.}
Overall, \algname{} keeps both costs moderate, supporting its scalability to longer-term dialogue.

\section{Conclusion}

In this work, we addressed a blind spot in long-term personalized dialogue: existing memory mechanisms organize past interactions well but leave the user's evolving persona detached from the events that ground it. We traced this to two fundamental gaps---the \emph{memory--persona validity gap} and the \emph{persona-aware retrieval gap}.
To close these gaps, we proposed \algname{}, an evidence-guided heterogeneous persona-memory graph that connects event- and persona-level nodes through typed evidence edges.
Rather than accumulating personas as a static profile, \algname{} keeps every persona signal traceable to the evidence that supports or revises it, and at retrieval time expands from query-relevant seeds along evidence edges to recover persona-critical evidence that direct query--memory similarity misses. Across three long-term personalized dialogue benchmarks, \algname{} consistently outperforms full-history, summary-based, persona-aware, graph-structured, and agentic memory baselines. 
We believe that organizing persona memory as an evidence graph offers a practical path toward reliable personalization, especially for resource-constrained SLM backbones, where personalization must be carried by memory structure rather than backbone capacity.

\smallskip
\section*{Limitations}
Although \algname{} demonstrates the value of relational graph structure for persona-aware memory modules, several limitations remain. 
The current evidence edge taxonomy captures core evidential polarities, but not finer-grained cognitive dimensions. Grounding it in cognitive models of belief revision could enable more nuanced validity reasoning beyond the current binary sign composition~\citep{comet21}.

Our experiments are confined to synthetic datasets, since authentic long-term dialogues with organic persona evolution are rarely public and remain proprietary due to privacy constraints~\citep{msc22}. Constructing privacy-preserving benchmarks that better approximate real-world persona dynamics is an important direction for future evaluation.
\rev{Furthermore, interactions spanning months or years would accumulate far more nodes and edges than the contexts we evaluate, so efficient lifecycle policies for pruning, consolidation, and decay become necessary in these lifelong scenarios. 
User data rights raise the same requirement, particularly in sensitive domains such as healthcare. 
Because persona signals are stored as individual, human-readable State and Trait nodes rather than a condensed profile, a user can delete a specific signal without discarding the rest, and low validity and \textsc{Shift\_{to}} ($\rightsquigarrow$) supersession already identify stale and superseded nodes as candidates for automatic pruning. 
We expect these policies to be simple additions rather than a redesign and leave them to future work.}

\smallskip
\section*{Acknowledgments}
This work was supported by the Institute of Information \& Communications Technology Planning \& Evaluation (IITP) grants funded by the Korea government(MSIT) (No. RS-2026-25507543 (40\%), No. IITP-2026-RS-2025-02304828 (40\%), No. IITP-2026-RS-2020-II201819 (10\%), and No. RS-2026-25585074 (10\%)).

\clearpage
\bibliography{custom}

\clearpage
\appendix
\section*{Appendix}

\section{Detailed Experiment Settings}
\label{app:appendix_experiment}

\subsection{Datasets}
\label{app:datasets}

Table~\ref{tab:dataset_stats} summarizes the statistics of the three benchmarks under our evaluation setup.

\paragraph{ImplexConv.}
ImplexConv~\citep{implexconv25} studies \emph{implicit reasoning}, where persona-relevant evidence is semantically distant from the query rather than explicitly stated.
It provides \emph{opposed} and \emph{supportive} subsets, and we use the \emph{opposed} subset.
An opposed scenario introduces a situation that contradicts a previously established trait, so a correct answer must reason over conflicting and superseding evidence.
We evaluate on $300$ conversation instances, yielding $1{,}259$ QA queries.

\paragraph{PersonaMem.}
PersonaMem~\citep{personamem25} targets \emph{dynamic user profiling}: whether a model tracks how user preferences evolve across temporally ordered, multi-topic sessions, including preference evolution and recommendation in unseen scenarios. 
We use the $32$k- and $128$k-token history settings, which let us examine how each method behaves as conversational history accumulates.

\paragraph{PrefEval.}
PrefEval~\citep{prefeval25} evaluates whether LLMs infer, retain, and follow user preferences in long context. We adapt PrefEval into a multi-persona discrimination setting: we treat each preference--query pair as an independent single-session persona and concatenate $N$ such sessions into one evolving conversation. After appending the $k$-th session, we evaluate the queries associated with all sessions observed up to $k$. This cumulative querying protocol yields $N(N+1)/2$ QA queries.

\subsection{Baselines}
\label{app:baselines}

We compare \algname{} with six representative agent memory baselines and a full-history reference baseline.

\begin{table}[t]
\centering
\small
\begin{tabular}{lrrr}
\toprule
 & \textbf{Sessions} & \textbf{Turns/sess.} & \textbf{QAs} \\
\midrule
ImplexConv   & $300$ & $656.2$ & $1{,}259$ \\
PrefEval     & $100$ & $530$   & $5{,}050$ \\
P.Mem-32k    & $37$  & $90.9$  & $589$     \\
P.Mem-128k   & $60$  & $406.4$ & $2{,}727$ \\
\bottomrule
\end{tabular}
\caption{Dataset statistics under our evaluation setup. For ImplexConv,
\emph{Sessions} counts conversation instances; for PrefEval and PersonaMem,
concatenated persona-sessions and sessions per history, respectively.
PrefEval QAs exceed turns due to cumulative re-querying.}
\label{tab:dataset_stats}
\end{table}

\begin{table*}[t]
\centering
\small
\resizebox{\textwidth}{!}{
\setlength{\tabcolsep}{6pt}
\renewcommand{\arraystretch}{1.2}
\begin{tabular}{llll}
\toprule
\textbf{Model} & \textbf{Parameters} & \textbf{License} & \textbf{Public Link} \\
\midrule
Qwen3-1.7B       & 1.7B & Apache License 2.0 & \url{huggingface.co/Qwen/Qwen3-1.7B} \\
Gemma-3-1B        & 1B   & Gemma License      & \url{huggingface.co/google/gemma-3-1b-it} \\
Qwen3.5-4B       & 4B   & Apache License 2.0 & \url{huggingface.co/Qwen/Qwen3.5-4B} \\
Gemma-3-4B        & 4B   & Gemma License      & \url{huggingface.co/google/gemma-3-4b-it} \\
all-MiniLM-L6-v2 & 22M  & Apache License 2.0 & \url{huggingface.co/sentence-transformers/all-MiniLM-L6-v2} \\
\bottomrule
\end{tabular}}
\caption{Licenses and sources of the model artifacts used in our experiments. All artifacts are publicly available and used in a manner consistent with their intended research use.}
\label{tab:artifact_licenses}
\end{table*}

\begin{itemize}
\item \textbf{Full History.} This baseline directly provides the entire available dialogue history to the backbone LLM without an explicit memory module. We include it as a memory-free reference point, testing whether structured memory provides benefits beyond naively conditioning on all past interactions.
\item \textbf{MemoryBank}~\citep{memorybank24}. MemoryBank is a summary-based long-term memory framework that compresses past dialogue into memory states while maintaining a user portrait. It represents a compact memory approach that combines efficient history compression with user-level personalization.
\item \textbf{LD-Agent}~\citep{ldagent25}. A persona-aware personalized dialogue agent that pairs event memory with a dedicated persona module. Its persona module is modeled separately and captures agent and user personas jointly, making it our representative for explicitly persona-conditioned memory.
\item \textbf{THEANINE}~\citep{theanine25}. A temporal graph-structured memory model that links memories through typed relations and retrieves relevant memory timelines. It provides a graph-based baseline for modeling memory evolution over long-term interactions.
\item \textbf{A-MEM}~\citep{amem25}. An agentic note-based memory system whose memory units autonomously link and evolve over time. It is included as a recent and competitive instantiation of the agentic memory paradigm.
\rev{\item \textbf{SeCom}~\citep{secom25}. A segment-level memory framework that constructs memory at the conversation-segment granularity and denoises retrieval through prompt compression with an external module (LLMLingua-2). As it relies heavily on an external compression module, we report it as a separate additional baseline in Appendix~\ref{app:additional-baselines}.
\item \textbf{H$^2$Memory}~\citep{mempal26}. A hierarchical and heterogeneous memory framework for long-term personalized assistance. We include it as a recent structured-memory comparison in Appendix~\ref{app:additional-baselines}.}

\end{itemize}

\begin{table*}[t]
\centering
\small
\setlength{\tabcolsep}{3.5pt}
\renewcommand{\arraystretch}{1.15}
\resizebox{\textwidth}{!}{
\begin{tabular}{p{0.14\textwidth} p{0.18\textwidth} p{0.38\textwidth} ccccc}
\toprule
\multirow{2}{*}{\textbf{Dimension}} 
& \multirow{2}{*}{\textbf{Sub-dimension}} 
& \multirow{2}{*}{\textbf{Seed Questions}} 
& \multicolumn{5}{c}{\textbf{Provided Information}} \\
\cmidrule(lr){4-8}
& & & \textbf{Question} & \textbf{Gen. Ans.} & \textbf{Persona Factor} & \textbf{Ref. Conv.} & \textbf{GT Ans.} \\
\midrule
\makecell[l]{\textbf{Response}\\\textbf{Competence}}
& Question Addressing
& Does the generated answer directly address what the question is asking?
& \(\checkmark\) & \(\checkmark\) & -- & -- & \(\checkmark\) \\

\midrule
\multirow{3}{*}{\makecell[l]{\textbf{Persona}\\\textbf{Adaptation}}}
& Persona Recognition
& Is the implicit persona factor reflected in the generated answer?
& \(\checkmark\) & \(\checkmark\) & \(\checkmark\) & \(\checkmark\) & -- \\
\cmidrule(lr){2-8}

& Generic Distinctness
& Does the answer read as tailored to a specific individual rather than as a generic answer?
& \(\checkmark\) & \(\checkmark\) & -- & -- & -- \\
\cmidrule(lr){2-8}

& Substantive Integration
& Is the persona integrated into the answer's substantive content choices, beyond a surface-level mention?
& \(\checkmark\) & \(\checkmark\) & \(\checkmark\) & \(\checkmark\) & \(\checkmark\) \\
\bottomrule
\end{tabular}
}
\caption{ImplexConv LLM-as-a-judge dimensions, adapted from CheckEval~\citep{checkeval25}. A checkmark indicates that the corresponding information is provided to the judge prompt for that sub-dimension.}
\label{tab:implexconv_dims}
\end{table*}

\begin{table*}[t]
\centering
\small
\setlength{\tabcolsep}{3.5pt}
\renewcommand{\arraystretch}{1.15}
\resizebox{\textwidth}{!}{
\begin{tabular}{p{0.15\textwidth} @{\hspace{1pt}} p{0.19\textwidth} p{0.42\textwidth} ccc}
\toprule
\multirow{2}{*}{\textbf{Dimension}} 
& \multirow{2}{*}{\textbf{Sub-dimension}} 
& \multirow{2}{*}{\textbf{Seed Questions}} 
& \multicolumn{3}{c}{\textbf{Provided Information}} \\
\cmidrule(lr){4-6}
& & & \textbf{Preference} & \textbf{Question} & \textbf{Generated Answer} \\
\midrule
\makecell[l]{\textbf{Preference}\\\textbf{Consistency}}
& \makecell[l]{Preference \\ Violation}
& Does the response avoid recommendations that violate the user's stated preference?
& \(\checkmark\) & \(\checkmark\) & \(\checkmark\) \\

\midrule
\makecell[l]{\textbf{Preference}\\\textbf{Awareness}}
& \makecell[l]{Preference \\ Acknowledgement}
& Does the response explicitly or implicitly acknowledge a user preference when answering the query?
& -- & \(\checkmark\) & \(\checkmark\) \\

\midrule
\makecell[l]{\textbf{Preference}\\\textbf{Grounding}}
& \makecell[l]{Preference \\ Hallucination}
& Does the response avoid misstating, contradicting, or fabricating the user's stated preference?
& \(\checkmark\) & -- & \(\checkmark\) \\

\midrule
\makecell[l]{\textbf{Response}\\\textbf{Utility}}
& Helpful Response
& Does the response provide substantive, query-relevant help without merely apologizing or asking for more information?
& -- & \(\checkmark\) & \(\checkmark\) \\
\bottomrule
\end{tabular}
}
\caption{PrefEval LLM-as-a-judge criteria, following \citet{prefeval25}. A checkmark indicates that the corresponding information is provided to the judge prompt. Scores are normalized so that higher values indicate better preference following.}
\label{tab:prefeval_dims}
\end{table*}

\subsection{Implementation Details}
\label{app:implementation}

\paragraph{Backbone LLMs.}
All methods use Qwen3-1.7B~\citep{qwen3_25} and Gemma-3-1B~\citep{gemma3_25} as SLM backbones.
As personalized agents move toward on-device and privacy-preserving deployment, SLMs are the practical backbone.
Their limited context leaves little room to recover an evolving persona from raw dialogue alone, so any gain reflects the memory structure rather than backbone capacity.
In our additional experiments (\cref{app:additional}), we further evaluate \algname{} with the larger Qwen3.5-4B~\citep{qwen3_25} and Gemma-3-4B~\citep{gemma3_25} backbones.

\paragraph{Embedding model.}
For all embedding-based operations such as seed retrieval, semantic similarity scoring, and embedding-based evaluation, we use all-MiniLM-L6-v2 as the sentence encoder. 
The same encoder is applied identically across \algname{} and all baselines.

All backbone LLMs and the embedding model are publicly released research artifacts. Their licenses and sources are listed in Table~\ref{tab:artifact_licenses}, and our use is consistent with their intended research use.

\paragraph{Temporal normalization.}
None of the three benchmarks defines an explicit temporal axis over its conversations, \rev{though the relative order of persona signals is native to all of them}.
Since several methods consume timestamps during memory construction and retrieval, we impose one normalization scheme applied uniformly to all methods.
We map one unit of conversational structure to a calendar day, namely a conversation cluster for ImplexConv, a system-prompt-delimited segment for PersonaMem, and a session for PrefEval.
Within a day, every turn is assigned a fixed 10-minute increment. 
Absolute values are immaterial, and the scheme is identical for every method, so it introduces no relative bias.
\rev{PGMem reads these timestamps only as an ordering. A \textsc{Shift\_To} edge encodes the relative relation $v_i \rightsquigarrow v_j$ (\cref{sec:edges}), and the signed traversal (\cref{sec:graph_expansion}) and validity score (\cref{eq:validity-score}) are defined over these relations rather than absolute time.
Genuinely timestamped histories would additionally support elapsed-time signals such as recency decay, which the order-based sign composition does not model.}

\paragraph{Hardware.}
All experiments are run on a heterogeneous GPU cluster of four node types, namely NVIDIA A100 (80 GB), A6000 (48 GB), RTX 4090 (24 GB), and RTX PRO 6000 Blackwell (96 GB), with four nodes each. \rev{Node type does not affect the reported scores. The latency measurements in Appendix~\ref{app:latency} are the exception, so all compared methods are timed on a single NVIDIA H100 (80 GB) node.}

\subsection{Evaluation details}
\label{app:evaluation}

\paragraph{LLM-as-a-judge.}
We evaluate model outputs with an LLM-as-a-judge for ImplexConv and PrefEval, as both benchmarks require free-form generation.
For ImplexConv, we adopt CheckEval~\citep{checkeval25}, \rev{which reports strong agreement with human judgments in its original evaluation ($\rho \approx 0.72$)} and decomposes evaluation into a binary checklist: each response is scored along two dimensions---response competence and persona adaptation---comprising four binary sub-dimensions in total (Table~\ref{tab:implexconv_dims}). 
The full ImplexConv judge prompts are provided in Figures~\ref{fig:judge_question_addressing}--\ref{fig:judge_substantive_integration}.
For PrefEval, we follow the four-criterion judge protocol of \citet{prefeval25}---violation, acknowledgment, hallucination, and helpfulness---each a binary check (Table~\ref{tab:prefeval_dims}). 
Every sub-dimension is judged independently and normalized so that $1$ denotes the desirable outcome; the per-response score is the sum of its checks, giving a $0$--$4$ scale for both benchmarks. 
We use gpt-4o-mini as the judge model with the temperature set to $0$.

\paragraph{Exact-match accuracy.}
PersonaMem~\citep{personamem25} casts each query as a four-way multiple-choice question with a single correct option. We evaluate it with exact-match accuracy.

\section{Details of \algname{}}
\label{app:method_details}

\subsection{Algorithms}
\label{app:algorithm}
We provide the detailed procedures for memory construction (\cref{sec:construction}) and memory retrieval (\cref{sec:retrieval}) in Algorithm~\ref{alg:construction} and Algorithm~\ref{alg:retrieval}, respectively.

\subsection{Hyperparameters}
\label{app:hyperparameters}

We report hyperparameters referenced but not fully specified in the main text; all values are shared across the three benchmarks and both backbones. 
For the seed score $\phi(q,v)$ in \cref{sec:seed_retrieval}, Context nodes use a fixed weight pair $(w_{\mathrm{sem}},w_{\mathrm{ov}})=(0.65,0.35)$, while
State, Episode, and Trait nodes use a scope-dependent pair: $(0.60,0.40)$ for \textsc{Narrow} and $(0.85,0.15)$ for \textsc{Broad}, down-weighting lexical overlap for broader-scope nodes. 
The base retrieval budget uses per-type seed counts $k_{\mathrm{seed}}=20,6,17,6$ for Context, Episode, State, and Trait nodes with $k_{\mathrm{aps}}=6$ APS slots, and final assembly keeps $5$ traits, $18$ states, and $4$ episodes, chosen so that the total number of retrieved memories is comparable to that of the other memory baselines. Multi-hop sign propagation is bounded by a hop cap $H=10$.

\begin{algorithm}[t]
\small
\caption{Memory Graph Construction.}
\label{alg:construction}
\begin{algorithmic}[1]
\Require utterance stream $\{u_t^{\rho_t}\}$, chunk size $C$, trait interval $B$
\Ensure  persona-memory graph $\mathcal{G}=(\mathcal{V},\mathcal{E})$

\State $\mathcal{G}\gets(\emptyset,\emptyset)$
\For{each dialogue turn $t$}
  \LComment{Event and persona node extraction}
  \State $v^{c}\gets\textsc{Context}(u_t^{\rho_t})$;\;
         add $v^{c}$ to $\mathcal{V}^{c}$
  \State $v^{s}\gets\textsc{ExtractState}(u_t^{\rho_t})$
  \If{$v^{s}\neq\varnothing$}
    \State add $v^{s}$ to $\mathcal{V}^{s}$;\;
           add $v^{c}\twoheadrightarrow v^{s}$ to $\mathcal{E}_{\mathrm{src}}$
    \State $\mathcal{E}_{\mathrm{evi}}\gets\mathcal{E}_{\mathrm{evi}}\cup
           \textsc{Judge}\big(\textsc{Pairs}(v^{s},\,\mathcal{V})\big)$
  \EndIf
  \LComment{Chunk boundary: episode summary}
  \If{$t$ closes a chunk}
    \State $v^{e}\gets\textsc{ExtractEpisode}(\text{chunk contexts})$
    \State add $v^{e}$ to $\mathcal{V}^{e}$;\;
           add $\{v^{c}\twoheadrightarrow v^{e}\}$ to $\mathcal{E}_{\mathrm{src}}$
    \State $\mathcal{E}_{\mathrm{evi}}\gets\mathcal{E}_{\mathrm{evi}}\cup
           \textsc{Judge}\big(\textsc{Pairs}(v^{e},\,\mathcal{V})\big)$
  \EndIf
  \LComment{Every $B$ chunks: trait abstraction}
  \If{$B$ chunks elapsed}
    \State $v^{t}\gets\textsc{ExtractTrait}
           (\text{recent }\mathcal{V}^{c},\mathcal{V}^{s},\mathcal{V}^{e})$
    \If{$v^{t}\neq\varnothing$}
      \State add $v^{t}$ to $\mathcal{V}^{t}$;\;
             add $\{v^{c},v^{s},v^{e}\}\twoheadrightarrow v^{t}$
             to $\mathcal{E}_{\mathrm{src}}$
      \State $\mathcal{E}_{\mathrm{evi}}\gets\mathcal{E}_{\mathrm{evi}}\cup
             \textsc{Judge}\big(\textsc{Pairs}(v^{t},\,\mathcal{V})\big)$
    \EndIf
  \EndIf
  \LComment{Offline update: long-range relations}
  \If{$B$ chunks elapsed}
    \State $\mathcal{E}_{\mathrm{evi}}\gets\mathcal{E}_{\mathrm{evi}}\cup
           \textsc{Judge}\big(\textsc{DistantPairs}(\mathcal{V})\big)$
  \EndIf
\EndFor
\State \Return $\mathcal{G}$
\end{algorithmic}
\end{algorithm}

\begin{algorithm}[t]
\small
\caption{Memory Retrieval.}
\label{alg:retrieval}
\begin{algorithmic}[1]
\Require query $q$, graph $\mathcal{G}=(\mathcal{V},\mathcal{E})$,
         hop cap $H$
\Ensure  validity-ranked evidence set $\mathcal{F}$

\LComment{Stage 1: Seed retrieval}
\State $\mathcal{W}_0\gets\bigcup_{\tau\in\{c,e,s,t\}}
       \textsc{TopK}_{\tau}\big(\phi(q,\cdot)\big)
       \;\cup\;\textsc{APS}(q)$
\Statex

\LComment{Stage 2: Evidence-guided expansion}
\State replace each $v^{c}\in\mathcal{W}_0$
       by its Source-linked $e,s,t$ nodes
\State $\mathcal{O}\gets\mathcal{W}_0\setminus
       \textsc{APS}(q)$
       \Comment{expansion origins}
\State $\mathcal{V}^{\mathrm{exp}}\gets\emptyset$
\For{each origin $o\in\mathcal{O}$}
  \For{each path from $o$ to $v$ within $H$ hops}
    \State $\sigma\gets$ composed sign along the path
    \If{$\sigma=s^{+}$}
      \State add $v$ to $\mathcal{V}^{\mathrm{exp}}$; continue traversal
    \Else
      \State add $v$ to $\mathcal{V}^{\mathrm{exp}}$; terminate path
    \EndIf
  \EndFor
\EndFor
\State $\mathcal{W}_1\gets\textsc{Dedup}
       (\mathcal{W}_0\cup\mathcal{V}^{\mathrm{exp}})$
\Statex

\LComment{Stage 3: Validity-aware assembly}
\State drop superseded ($\rightsquigarrow$) sources
       whose targets are in $\mathcal{W}_1$
\For{each $v\in\mathcal{W}_1$}
  \State $\mathrm{val}(v)\gets
         \dfrac{|\mathcal{S}(v)|+\alpha}
               {|\mathcal{S}(v)|+|\mathcal{C}(v)|+2\alpha}$
  \State $\psi(q, v) \leftarrow \lambda\, \mathrm{val}(v) + (1 - \lambda)\, \phi(q, v)$
\EndFor
\State $\mathcal{F}\gets$ top-ranked nodes per type under $\psi$,
       serialized by type and validity status
\State \Return $\mathcal{F}$
\end{algorithmic}
\end{algorithm}

\subsection{Prompt}
\label{app:prompt}

Figure~\ref{fig:shared_prompt_components} summarizes the shared prompt components reused across extraction, relation classification, and QA generation.
Figures~\ref{fig:state_extraction_prompt}--\ref{fig:trait_extraction_prompt} present the prompts for extracting State, Episode, and Trait nodes.
Figure~\ref{fig:trait_state_relation_prompt} presents the evidence relation classification prompt, which determines whether two memory nodes support, contradict, shift to, or are irrelevant to each other.
Although the figure illustrates the Trait-centered case, a similar prompting scheme is applied to other State--Episode--Trait pairs by changing the input node types and direction rules.
Figure~\ref{fig:qa_opposed_prompt} presents the ImplexConv QA prompt, where retrieved States, Traits, Episodes, and Challenged Traits are used as candidate evidence for personalized answer generation.

\section{Detailed Experimental Results}
\label{app:results}

\begin{table}[t]
\centering
\setlength{\tabcolsep}{3pt}
\renewcommand{\arraystretch}{1.15}
\resizebox{\columnwidth}{!}{%
\begin{tabular}{llccccc}
\toprule
\textbf{Base LLM} & \textbf{Method} & \textbf{Total}
& \textbf{RC-QA} & \textbf{PA-Rec} & \textbf{PA-Dist} & \textbf{PA-Int} \\
\midrule
\multirow{6}{*}{\textbf{Qwen3-1.7B}}
& Full History & 0.99 & 0.96 & 0.01 & 0.00 & 0.01 \\
& MemoryBank   & 1.03 & 0.96 & 0.03 & 0.01 & 0.03 \\
& LD-Agent     & 0.92 & 0.89 & 0.01 & 0.01 & 0.01 \\
& A-MEM        & 0.94 & 0.91 & 0.01 & 0.00 & 0.02 \\
& THEANINE     & 0.98 & 0.95 & 0.01 & 0.01 & 0.02 \\
\cmidrule(lr){2-7}
& \cellcolor{oursrowlight}\textbf{\algname{}}   
& \cellcolor{oursrowlight}1.33 
& \cellcolor{oursrowlight}0.95 
& \cellcolor{oursrowlight}0.02 
& \cellcolor{oursrowlight}0.29 
& \cellcolor{oursrowlight}0.06 \\
\midrule
\multirow{6}{*}{\textbf{Gemma-3-1B}}
& Full History & 0.86 & 0.80 & 0.01 & 0.02 & 0.03 \\
& MemoryBank   & 1.09 & 0.91 & 0.04 & 0.05 & 0.09 \\
& LD-Agent     & 0.89 & 0.80 & 0.01 & 0.07 & 0.02 \\
& A-MEM        & 0.57 & 0.48 & 0.01 & 0.06 & 0.02 \\
& THEANINE     & 0.98 & 0.87 & 0.02 & 0.07 & 0.03 \\
\cmidrule(lr){2-7}
& \cellcolor{oursrowlight}\textbf{\algname{}}   
& \cellcolor{oursrowlight}1.41 
& \cellcolor{oursrowlight}0.77 
& \cellcolor{oursrowlight}0.03 
& \cellcolor{oursrowlight}0.52 
& \cellcolor{oursrowlight}0.09 \\
\bottomrule
\end{tabular}}
\caption{Per-dimension LLM-judge scores on ImplexConv. RC-QA:
question addressing; PA-Rec: persona recognition; PA-Dist: generic
distinctness; PA-Int: substantive integration.}
\label{tab:llmjudge-implexconv}
\end{table}

\begin{table}[t]
\centering
\setlength{\tabcolsep}{3pt}
\renewcommand{\arraystretch}{1.15}
\resizebox{\columnwidth}{!}{%
\begin{tabular}{llccccc}
\toprule
\textbf{Base LLM} & \textbf{Method} & \textbf{Total}
& \textbf{Viol.} & \textbf{Ack.} & \textbf{Halluc.} & \textbf{Helpful} \\
\midrule
\multirow{6}{*}{\textbf{Qwen3-1.7B}}
& Full History & 1.32 & 0.14 & 0.26 & 0.00 & 0.91 \\
& MemoryBank   & 1.21 & 0.05 & 0.16 & 0.00 & 1.00 \\
& LD-Agent     & 1.14 & 0.05 & 0.10 & 0.00 & 0.99 \\
& A-MEM        & 1.39 & 0.19 & 0.23 & 0.00 & 0.97 \\
& THEANINE     & 1.25 & 0.06 & 0.18 & 0.00 & 1.00 \\
\cmidrule(lr){2-7}
& \cellcolor{oursrowlight}\textbf{\algname{}}   
& \cellcolor{oursrowlight}2.21 
& \cellcolor{oursrowlight}0.25 
& \cellcolor{oursrowlight}0.94 
& \cellcolor{oursrowlight}0.02 
& \cellcolor{oursrowlight}1.00 \\
\midrule
\multirow{6}{*}{\textbf{Gemma-3-1B}}
& Full History & 1.44 & 0.06 & 0.40 & 0.00 & 0.97 \\
& MemoryBank   & 1.79 & 0.05 & 0.75 & 0.00 & 0.99 \\
& LD-Agent     & 1.61 & 0.07 & 0.55 & 0.00 & 0.99 \\
& A-MEM        & 1.60 & 0.21 & 0.57 & 0.00 & 0.82 \\
& THEANINE     & 1.55 & 0.05 & 0.51 & 0.00 & 1.00 \\
\cmidrule(lr){2-7}
& \cellcolor{oursrowlight}\textbf{\algname{}}   
& \cellcolor{oursrowlight}2.03 
& \cellcolor{oursrowlight}0.07 
& \cellcolor{oursrowlight}0.98 
& \cellcolor{oursrowlight}0.00 
& \cellcolor{oursrowlight}0.98 \\
\bottomrule
\end{tabular}}
\caption{Per-dimension LLM-judge scores on PrefEval. Viol.: preference violation; Ack.: preference acknowledgement; Halluc.: hallucination; Helpful: helpfulness.}
\label{tab:llmjudge-prefeval}
\end{table}

\subsection{LLM-as-a-judge Results}
\label{app:llm_judge_results}

This subsection reports the per-dimension breakdown of the LLM-judge scores summarized in the main results (Table~\ref{tab:main-results}).
Table~\ref{tab:llmjudge-implexconv} and Table~\ref{tab:llmjudge-prefeval} give the dimension-level scores for ImplexConv and PrefEval, respectively; the definition of each dimension follows the evaluation protocol described in \cref{app:evaluation}.

On ImplexConv (Table~\ref{tab:llmjudge-implexconv}), all methods reach comparable response-competence scores (\emph{rc\_question\_addressing}), while diverging sharply on the persona-adaptation dimensions: \algname{} attains the highest \emph{pa\_generic\_distinctness} and \emph{pa\_substantive\_integration}, indicating that its gains come from persona-aware retrieval rather than general answer quality. On PrefEval (Table~\ref{tab:llmjudge-prefeval}), \algname{} reaches the highest persona acknowledgement under both backbones ($0.94$ and $0.98$) while keeping helpfulness on par with baselines.

\subsection{Human Validation Results}
\label{app:human_validation}

\begin{table}[t]
\centering
\small
\setlength{\tabcolsep}{3pt}
\renewcommand{\arraystretch}{1.15}
\begin{tabular}{lccccc}
\toprule
\textbf{Method} & RC-QA & PA-Rec & PA-Dist & PA-Int & \textbf{Score} \\
\midrule
MemoryBank          & 1.00 & 0.27 & 0.37 & 0.40 & 2.03 \\
A-MEM               & 1.00 & 0.13 & 0.10 & 0.23 & 1.47 \\
\rowcolor{oursrowlight}
\textbf{\algname{}} & 1.00 & 0.23 & 0.80 & 0.23 & \textbf{2.27} \\
\bottomrule
\end{tabular}
\caption{Human validation results on 30 ImplexConv samples. Dimensions follow the binary checklist of Table~\ref{tab:implexconv_dims}.}
\label{tab:human_eval}
\end{table}

We conduct a human validation on ImplexConv to verify that our LLM-as-a-judge scores faithfully reflect human judgment. Two of the authors independently score the responses of \algname{}, MemoryBank, and A-MEM on 30 sampled queries, applying the same four-dimension checklist used by the LLM judge (\cref{app:evaluation}). Results are reported in Table~\ref{tab:human_eval}.

The human evaluation yields the same method ranking as the LLM judge, with \algname{} scoring
highest. The per-dimension pattern is also consistent. Question addressing is saturated across
all methods, so the separation arises almost entirely from the persona-adaptation dimensions,
on which \algname{} attains the highest score. This agreement indicates that the LLM-judge scores reported in the main results faithfully reflect human judgment.

\section{Analysis of Memory Construction}
\subsection{LLM Calls and Latency}
\label{app:latency}

\rev{
Table~\ref{tab:scaling} measures efficiency in tokens.
We complement it with the number of LLM calls and the wall-clock latency of memory construction.
We compare A-MEM, THEANINE, and \algname{}, all of which construct memory through repeated LLM calls.
Table~\ref{tab:latency} reports the results on ImplexConv with Qwen3-1.7B, measured on a single NVIDIA H100 (80\,GB) node.

Construction is incremental, and each utterance updates the graph as the dialogue unfolds (Section~\ref{sec:construction}).
\algname{} spends 1,441\,s over an average 656.2-turn session (Table~\ref{tab:dataset_stats}), which is 2.2\,s per turn.
A-MEM spends 2.4\,s per turn under the same accounting.
\algname{} also issues fewer construction calls than A-MEM, 1,804 vs 2,596, or 2.8 vs 4.0 per turn.
THEANINE is the cheapest to construct at 341\,s.
At query time, \algname{} and A-MEM issue a single generation call.
THEANINE issues 6 calls per query for timeline refinement.
Read together with the QA token counts in Table~\ref{tab:scaling}, \algname{} is lighter than A-MEM in tokens (1,847 vs 5,124) and lighter than THEANINE in calls (1 vs 6).
}

\begin{table}[t]
\centering
\small
\setlength{\tabcolsep}{5pt}
\begin{tabular}{lrrr}
\toprule
\multirow{2}{*}{Method} & \multicolumn{2}{c}{Construction} & QA \\
\cmidrule(lr){2-3}\cmidrule(lr){4-4}
 & Calls (\#) & Lat. (s) & Calls (\#) \\
\midrule
A-MEM & 2,596 & 1,584 & 1 \\
THEANINE & 1,254 & 341 & 6 \\
\algname{} & 1,804 & 1,441 & 1 \\
\bottomrule
\end{tabular}
\caption{LLM calls and latency on ImplexConv with Qwen3-1.7B. Construction is the total over a full session. QA Calls is the number of LLM calls per query at inference.}
\label{tab:latency}
\end{table}

\subsection{Reliability of Evidence Relation Labeling}
\label{app:edge-cases}

\rev{\algname{} performs on average 7.2k relation judgments per session, over 2.1M pairs across the 300 ImplexConv sessions.
This subsection examines what this labeling step produces and how reliable its output is.
Table~\ref{tab:edge-stats} reports per-session edge counts on ImplexConv with Qwen3-1.7B.
\textsc{Sup} ($\oplus$) edges dominate and consolidate a stable persona.
\textsc{Con} ($\ominus$) and \textsc{Sft} ($\rightsquigarrow$) edges are far rarer.
They connect opposing signals and temporal changes, capturing how the persona is challenged or revised over time.
\textsc{Irr} ($\emptyset$) absorbs pairs without clear evidential force, following the decision priority in Figure~\ref{fig:trait_state_relation_prompt}.

\paragraph{Human validation.}
We randomly sample 25 edges per relation type from the constructed graphs, 100 edges in total.
Two of the authors independently judge each edge against the label definitions of Figure~\ref{fig:trait_state_relation_prompt}.
Each judgment is binary, and we report accuracy as the percentage judged correct.
Accuracy is high on \textsc{Sup} ($\oplus$) and \textsc{Irr} ($\emptyset$) and low on \textsc{Con} ($\ominus$) and \textsc{Sft} ($\rightsquigarrow$).
Separating a genuine conflict or a genuine supersession from an unrelated pair requires tracking which persona signal is still active.
SLM backbones often fail at this, so the two relations that carry persona change are also the two they label least reliably.
The sample draws equally from the four types.
\textsc{Con} ($\ominus$) and \textsc{Sft} ($\rightsquigarrow$) are far rarer in the graph than the other two, so their accuracies apply to a small portion of all edges.

\paragraph{Representative cases.}
We show correct and mislabeled edges for the two hardest relation types ([s] State, [t] Trait, [e] Episode).
\begin{itemize}
\item \textit{Correct \textsc{Con}.} [s] struggling to stay motivated during workouts $\overset{\ominus}{\longrightarrow}$ [t] a dedicated, motivated individual committed to consistent, progressive training. The two co-active signals are in direct tension.
\item \textit{Correct \textsc{Sft}.} [s] interested in golf, has never tried it $\rightsquigarrow$ [s] has participated in local tournaments and wants to use their skills for a greater purpose. The newer signal supersedes the older one.
\item \textit{Mislabeled \textsc{Con}.} [e] overwhelmed by a family member's personal crisis $\overset{\ominus}{\longrightarrow}$ [t] highly concerned with finding quality educational apps for their kids. The pair carries no evidential tension, so the correct label is \textsc{Irr}.
\item \textit{Mislabeled \textsc{Sft}.} [s] overwhelmed today, needs someone to talk to $\rightsquigarrow$ [s] feels proud after reaching the top of the ladder. Two transient states are linked as an update although neither supersedes the other.
\end{itemize}
Both mislabels assign evidential force where none exists.
Removing \textsc{Con} ($\ominus$) or \textsc{Sft} ($\rightsquigarrow$) entirely costs 1.98 and 1.91 points on PersonaMem-128k (Table~\ref{tab:ablation}), which bounds what unreliable labeling can cost.
}

\begin{table}[t]
\centering
\small
\renewcommand{\arraystretch}{1.2}
\setlength{\tabcolsep}{3pt}
\begin{tabular}{lcccc}
\toprule
 & \textsc{Sup} ($\oplus$) & \textsc{Con} ($\ominus$) & \textsc{Sft} ($\rightsquigarrow$) & \textsc{Irr} ($\emptyset$) \\
\midrule
\makecell[l]{Edges \\ per session} & 5259.9 & 49.3 & 125.2 & 1722.9 \\
\makecell[l]{Human \\ validation (\%)} & 80.0 & 36.0 & 52.0 & 84.0 \\
\bottomrule
\end{tabular}
\caption{\rev{Per-session evidence edge statistics and human validation on ImplexConv with Qwen3-1.7B. Validation reports binary accuracy over 25 sampled edges per relation type.}}
\label{tab:edge-stats}
\end{table}

\section{Additional Experiments}
\label{app:additional}

\subsection{Additional Structured-Memory Baselines}
\label{app:additional-baselines}

\rev{We compare \algname{} with SeCom~\citep{secom25} and H$^2$Memory~\citep{mempal26} under the main SLM backbones, keeping all benchmarks, splits, retrieval budgets, and evaluation protocols identical to Section~\ref{sec:experiments}. 
Results are reported in Table~\ref{tab:additional-baselines}. 
\algname{} achieves the strongest results on ImplexConv and PersonaMem under both backbones. 
On PrefEval, \algname{} is strongest under Qwen3-1.7B, while H$^2$Memory and SeCom score slightly higher under Gemma-3-1B. 
Neither method links memory and persona entries through typed evidential relations, so persona signals superseded by later interactions are not down-weighted at retrieval time.}

\begin{table}[!t]
\centering
\small
\renewcommand{\arraystretch}{1.2}
\setlength{\tabcolsep}{3pt}
\begin{tabular}{llcccc}
\toprule
\multirow{2}{*}{} & \multirow{2}{*}{\textbf{Method}} 
& \multirow{2}{*}{\textbf{ImplexConv}} 
& \multirow{2}{*}{\textbf{PrefEval}} 
& \multicolumn{2}{c}{\textbf{PersonaMem}} \\ 
\cmidrule(lr){5-6}
& & & & 32k & 128k \\ 
\midrule
\multirow{3}{*}{\rotatebox{90}{\parbox{1.6cm}{\centering\textbf{Qwen3-1.7B}}}}
& H$^2$Memory  & \underline{0.90} & \underline{1.93} & 34.97 & 35.46 \\
& SeCom        & 0.83 & 1.81 & \underline{38.88} & \underline{37.15} \\
\cmidrule(lr){2-6}
& \cellcolor{oursrowlight}\textbf{\algname{}} 
& \cellcolor{oursrowlight}\textbf{1.33} 
& \cellcolor{oursrowlight}\textbf{2.21} 
& \cellcolor{oursrowlight}\textbf{45.50} 
& \cellcolor{oursrowlight}\textbf{46.35} \\
\midrule
\multirow{3}{*}{\rotatebox{90}{\parbox{1.6cm}{\centering\textbf{Gemma-3-1B}}}}
& H$^2$Memory  & \underline{0.98} & \textbf{2.12} & \underline{24.45} & \underline{29.08} \\
& SeCom        & 0.76 & \underline{2.06} & 24.11 & \underline{29.08} \\
\cmidrule(lr){2-6}
& \cellcolor{oursrowlight}\textbf{\algname{}} 
& \cellcolor{oursrowlight}\textbf{1.41} 
& \cellcolor{oursrowlight}2.03 
& \cellcolor{oursrowlight}\textbf{30.39} 
& \cellcolor{oursrowlight}\textbf{34.18} \\
\bottomrule
\end{tabular}
\caption{\label{tab:additional-baselines}
\rev{Comparison with additional structured-memory baselines under the main SLM backbones. 
Metrics follow Table~\ref{tab:main-results}. 
\textbf{Bold} and \underline{underline} denote the best and second-best results, respectively.}}
\end{table}

\subsection{Generalization to Larger Backbones}
\label{app:4b-experiments}
Our main experiments (Section~\ref{sec:experiments}) evaluate \algname{} under SLM backbones, where limited context length and reasoning capacity make an explicit persona-memory structure most impactful. 
We re-run \algname{} and all baselines with Qwen3.5-4B~\citep{qwen3_25} and Gemma-3-4B~\citep{gemma3_25}, keeping all benchmarks, splits, retrieval budgets, and evaluation protocols identical to the main experiments (\cref{app:appendix_experiment}). 
Results are reported in Table~\ref{tab:4b-results}.

\algname{} continues to achieve the strongest results on ImplexConv and PrefEval across both backbones, mirroring the ranking observed under SLMs. 
On PersonaMem, A-MEM instead obtains the highest accuracy.
We attribute this to its construction procedure, which depends substantially on the backbone model for memory organization and summarization and therefore benefits more directly from increased model capacity.
Consequently, A-MEM exhibits a larger relative gain at the 4B scale than under SLMs.
\rev{The gain is confined to this benchmark format. 
PersonaMem scores 4-way multiple-choice exact match (\cref{app:appendix_experiment}), so an item is answered by recalling a single past episode with no conflicting signal to reconcile. 
ImplexConv and PrefEval instead require superseded or conflicting signals to constrain a free-form response. 
Under the 4B backbones A-MEM scores 0.77 and 0.90 on ImplexConv with Qwen and Gemma, against 2.41 and 2.21 for \algname{}.}

\rev{\algname{} improves on every benchmark under both backbones when moving from the SLM to the 4B scale. 
With Qwen it rises from 1.33 to 2.41 on ImplexConv and from 45.50 to 62.65 on PersonaMem-32k (Tables~\ref{tab:main-results} and~\ref{tab:4b-results}).
Its relative advantage is nonetheless largest} in resource-constrained settings, where personalization is driven by memory structure rather than backbone capacity.
Refining how nodes, edges, and relations are extracted for stronger backbones is a natural extension, \rev{and we expect it to widen the margin at larger scales.}
We leave this direction to future work.

\begin{table}[!t]
\centering
\small
\renewcommand{\arraystretch}{1.1}
\setlength{\tabcolsep}{3pt}
\begin{tabular}{llcccc}
\toprule
\multirow{2}{*}{} & \multirow{2}{*}{\textbf{Method}} 
& \multirow{2}{*}{\textbf{ImplexConv}} 
& \multirow{2}{*}{\textbf{PrefEval}} 
& \multicolumn{2}{c}{\textbf{PersonaMem}} \\ 
\cmidrule(lr){5-6}
& & & & 32k & 128k \\ 
\midrule
\multirow{6}{*}{\rotatebox{90}{\textbf{Qwen3.5-4B}}} 
& Full history & 0.46 & 1.00 & \underline{65.37} & 46.17 \\
& MemoryBank   & \underline{2.31} & 2.40 & 61.12 & 49.28 \\
& LD-Agent     & 0.65 & 1.90 & 53.14 & 39.79 \\
& A-MEM        & 0.77 & 1.74 & \textbf{68.42} & \textbf{61.57} \\
& THEANINE     & 1.29 & \underline{2.46} & 60.61 & 52.73 \\
\cmidrule(lr){2-6}
& \cellcolor{oursrowlight}\textbf{\algname{}}  
& \cellcolor{oursrowlight}\textbf{2.41} 
& \cellcolor{oursrowlight}\textbf{2.57} 
& \cellcolor{oursrowlight}62.65 
& \cellcolor{oursrowlight}\underline{56.07} \\ 
\midrule
\multirow{6}{*}{\rotatebox{90}{\textbf{Gemma-3-4B}}}
& Full history & 1.12 & 1.96 & \underline{52.97} & 46.75 \\
& MemoryBank   & \underline{2.10} & 1.92 & 50.25 & 47.08 \\
& LD-Agent     & 1.09 & 2.11 & 49.58 & 45.32 \\
& A-MEM        & 0.90 & 1.93 & \textbf{53.48} & \textbf{53.21} \\
& THEANINE     & 1.59 & \underline{2.15} & 52.63 & 47.05 \\
\cmidrule(lr){2-6}
& \cellcolor{oursrowlight}\textbf{\algname{}}  
& \cellcolor{oursrowlight}\textbf{2.21} 
& \cellcolor{oursrowlight}\textbf{2.21} 
& \cellcolor{oursrowlight}51.78 
& \cellcolor{oursrowlight}\underline{49.21} \\ 
\bottomrule
\end{tabular}
\caption{\label{tab:4b-results}
Comparison results on long-term personalized dialogue benchmarks under larger backbone LLMs.
ImplexConv and PrefEval use LLM-as-a-judge scores for personalized response quality on a 5-point scale, while PersonaMem evaluates user persona QAs with accuracy (\%).
\textbf{Bold} and \underline{underline} denote the best and second-best results, respectively.}
\end{table}

\section{Case Study}

\rev{We present case studies of \algname{}, covering both success and failure cases.}

\label{app:case_study}

\begin{figure*}[t]
  \centering
  \includegraphics[width=\textwidth]{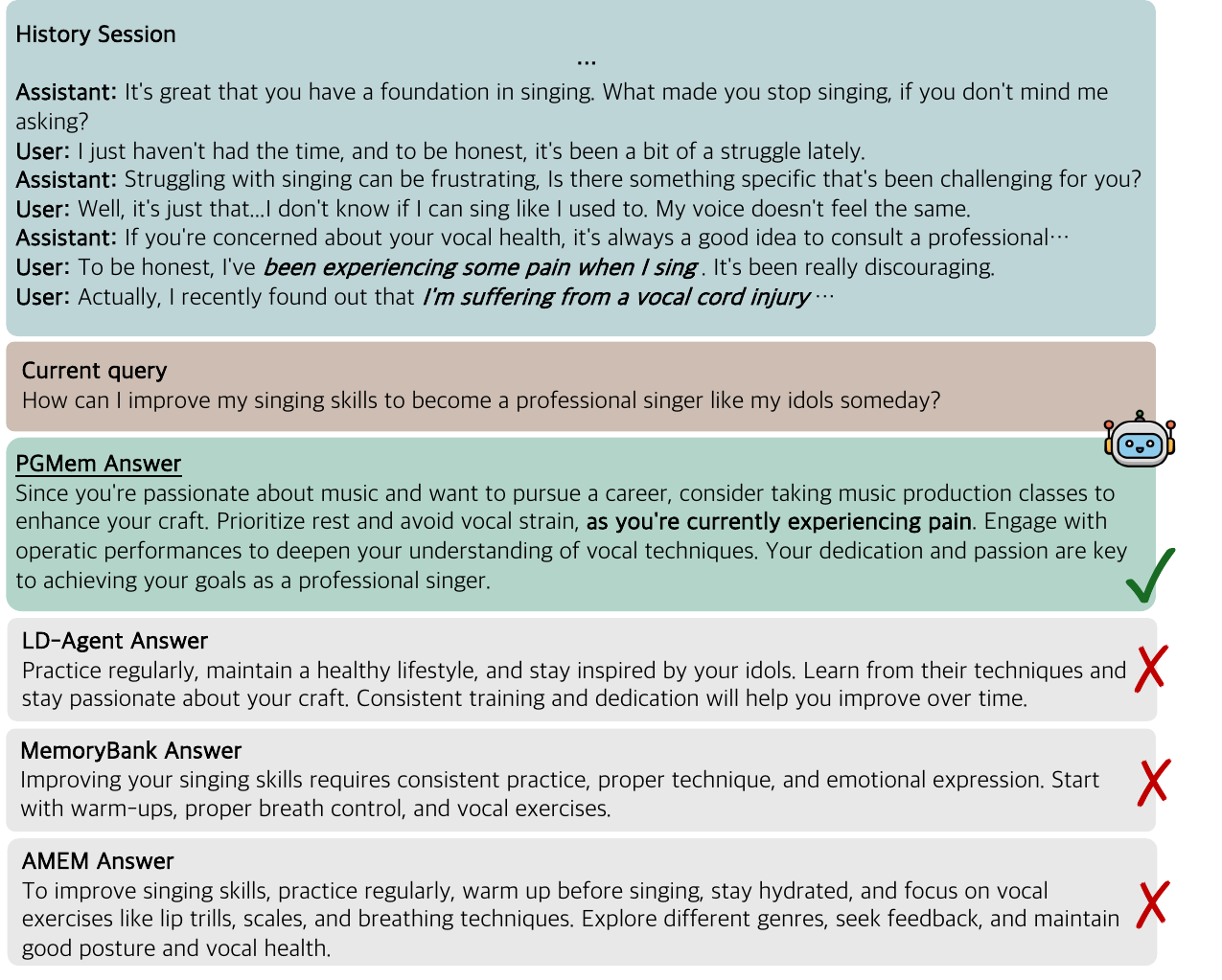}
  \caption{\rev{Success case} where the persona-critical signal, a vocal cord injury, is semantically distant from the query about becoming a professional singer. \rev{Baselines return generic or harmful advice, while \algname{} recovers the signal through evidence-guided expansion.}}
  \label{fig:case1}
\end{figure*}

\begin{figure*}[t]
\centering
\includegraphics[width=\textwidth]{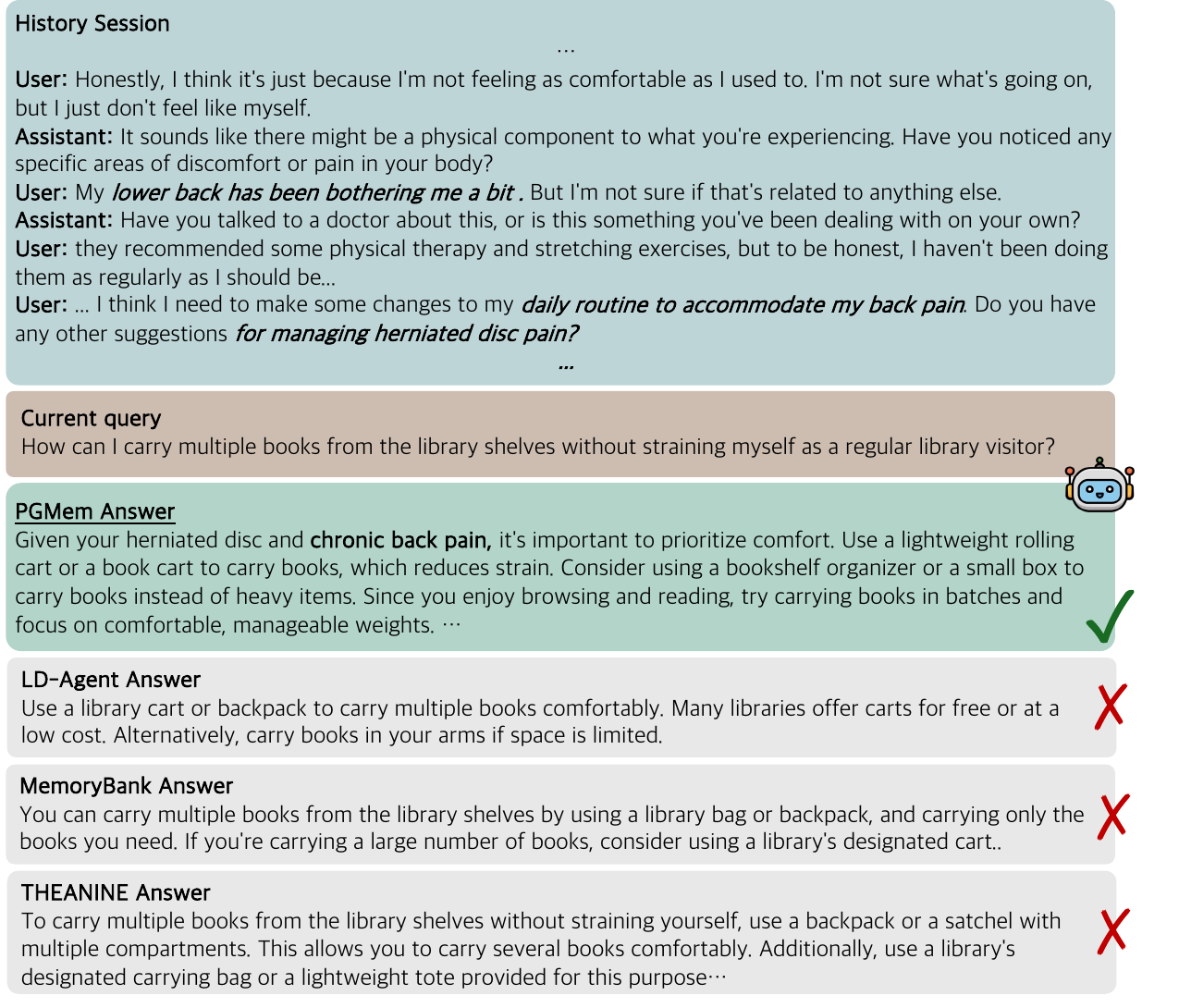}
\caption{\rev{Success case} where the persona-critical signal, a herniated disc, is semantically distant from the query about carrying library books. \rev{Baselines overlook the injury, while \algname{} surfaces it and adapts the answer.}}
\label{fig:case2}
\end{figure*}

\rev{\subsection{Success Cases}
\label{app:case_success}}
We \rev{first} inspect two cases where the query is topically far from the persona signal the correct answer depends on.
In Figures~\ref{fig:case1} and~\ref{fig:case2}, the baselines miss the user's vocal cord injury and herniated disc, returning generic or even harmful advice.
\algname{} expands along evidence edges, recovers these State nodes, and generates personalized answers.

\rev{\subsection{Failure Cases}
\label{app:case_failure}
We observe two failure modes, with one example of each. 
In near-miss cases the persona signal is retrieved but not reflected in substance. 
In complete-failure cases neither seed retrieval nor graph expansion surfaces the signal.

\paragraph{Near-miss.}
This case arises from the query ``How can I maintain a flawless complexion as a model for upcoming photoshoots and runway events?''
\begin{itemize}[leftmargin=*]
\setlength{\itemsep}{0pt}
\item \textbf{Retrieved memory.} [s] acne and dark spots. [s] visiting dermatologists for skin conditions. [s] persistent skin issues with no long-term solution. [e] low self-confidence related to skin issues.
\item \textbf{Generated.} ``...consistent skincare routine... address your acne and dark spots by consulting dermatologists... maintain a polished appearance.''
\item \textbf{Gold.} ``...a flawless complexion can be challenging due to a skin condition causing acne and hyperpigmentation... prioritizing skin health over flawless appearance can be a vital approach.''
\end{itemize}
The signal is retrieved and even mentioned, but the substantive stance is not adopted.
The response keeps a flawless-is-achievable framing while the gold reframes the goal toward skin health.
This is a generation-side limitation rather than a retrieval failure.

\paragraph{Complete-failure.}
This case arises from the query ``What steps should I take to secure a car loan for buying my first car?''
\begin{itemize}[leftmargin=*]
\setlength{\itemsep}{0pt}
\item \textbf{Retrieved memory.} No node about the user's credit history is surfaced.
\item \textbf{Generated.} ``...compare interest rates. Ensure you have a good credit score and a stable income...''
\item \textbf{Gold.} ``...to secure a car loan despite a low credit score caused by missed student-loan payments, consider improving your credit first... immediate approval may be challenging.''
\end{itemize}
The persona signal is not recovered, so generation defaults to generic advice assuming good credit.
}

\section{AI assistant usage acknowledgments}
Generative AI (Claude, ChatGPT) was used solely for grammar correction and language polishing of the manuscript. All content, analyses, results, and conclusions were developed and verified by the authors.

\begin{figure*}[p]
    \centering
    \begin{tcolorbox}[
        colback=gray!10!white, colframe=black, colbacktitle=gray!50!white,
        coltitle=black, rounded corners, boxrule=0.6mm, width=0.92\textwidth,
        fonttitle=\normalsize, fontupper=\footnotesize,
        title={\textbf{Question Addressing Judge Prompt (Response Competence)}},
    ]
        \textit{Your task is to evaluate one specific aspect of response competence at a time.}
        
        \medskip
        \textbf{\# Task Description}\\
        A dialogue system has generated an answer to a user's question. \\
        Your job: judge \textbf{whether the generated answer directly addresses what the question is asking}.

        \medskip
        \textbf{\# Evaluation Target}\\
        Question:~\texttt{\{query\}}\\
        Reference Answer:~\texttt{\{gt\_answer\}}\\
        Generated Answer:~\texttt{\{generated\_answer\}}

        \medskip
        \textbf{\# Scoring Rubric --- Question Addressing (0 / 1)}
        \begin{itemize}[nosep,leftmargin=*]
            \item \textbf{1} --- Addressed: The generated answer directly responds to what the question asks. It engages with the actual request rather than deflecting, changing topic, or restating the question.
            \item \textbf{0} --- Not Addressed: The answer is off-topic, fails to engage with the question, or provides a response so general it could apply to any question.
        \end{itemize}

        \medskip
        \textbf{Answer:}
    \end{tcolorbox}
    \captionof{figure}{Judge prompt for the \emph{Question Addressing} sub-dimension (Response competence).}
    \label{fig:judge_question_addressing}
    \vspace{5mm}

    \begin{tcolorbox}[
        colback=gray!10!white, colframe=black, colbacktitle=gray!50!white,
        coltitle=black, rounded corners, boxrule=0.6mm, width=0.92\textwidth,
        fonttitle=\normalsize, fontupper=\footnotesize,
        title={\textbf{Persona Recognition Judge Prompt (Persona Adaptation)}},
    ]
        \textit{You are an expert evaluator for a personalized conversational AI system. Your task is to assess one specific aspect of persona adaptation at a time. Focus only on the requested aspect.}

        \medskip
        \textbf{\# Task Description}\\
        A memory-augmented dialogue system has generated an answer for a user with a hidden \textbf{implicit persona factor} --- a state, preference, constraint, or circumstance that should shape the response.\\
        Your job: judge \textbf{whether the implicit persona factor is reflected in the generated answer}.

        \medskip
        \textbf{\# Implicit Persona Factor}\\
        \texttt{\{reason\}}

        \medskip
        \textbf{\# Supporting Evidence (relevant past conversations)}\\
        \texttt{\{reference\_conversation\}}

        \medskip
        \textbf{\# Evaluation Target}\\
        Question:~\texttt{\{query\}}\\
        Generated Answer:~\texttt{\{generated\_answer\}}

        \medskip
        \textbf{\# Scoring Rubric --- Persona Recognition (0 / 1)}
        \begin{itemize}[nosep,leftmargin=*]
            \item \textbf{1} --- Recognized: The generated answer either acknowledges the implicit persona factor or makes a clear response choice because of it.
            \item \textbf{0} --- Not Recognized: The answer shows no evidence that the persona factor motivated any content choice. It reads as a generic response to the question alone.
        \end{itemize}

        \medskip
        \textbf{Answer:}
    \end{tcolorbox}
    \captionof{figure}{Judge prompt for the \emph{Persona Recognition} sub-dimension (Persona adaptation).}
    \label{fig:judge_persona_recognition}
\end{figure*}

\begin{figure*}[p]
    \centering
    \begin{tcolorbox}[
        colback=gray!10!white, colframe=black, colbacktitle=gray!50!white,
        coltitle=black, rounded corners, boxrule=0.6mm, width=0.92\textwidth,
        fonttitle=\normalsize, fontupper=\footnotesize,
        title={\textbf{Generic Distinctness Judge Prompt (Persona Adaptation)}},
    ]
        \textit{You are an expert evaluator for a personalized conversational AI system. Your task is to assess one specific aspect of persona adaptation at a time. Focus only on the requested aspect.}

        \medskip
        \textbf{\# Task Description}\\
        A memory-augmented dialogue system has generated an answer for a user who has hidden persona context (state, preference, constraint, or circumstance not visible to you). \\
        Your job: judge whether the answer reads as \textbf{tailored to a specific individual} or as a \textbf{generic answer that could be given to anyone} asking the same question.

        \medskip
        \textbf{\# Evaluation Target}\\
        Question:~\texttt{\{query\}}\\
        Generated Answer:~\texttt{\{generated\_answer\}}

        \medskip
        \textbf{\# Scoring Rubric --- Generic Distinctness (0 / 1)}
        \begin{itemize}[nosep,leftmargin=*]
            \item \textbf{1} --- Distinct: The answer reads as personalized to a specific user. It references or builds on user-specific details (situation, preferences, constraints, prior choices, context), adapts its recommendation/framing to the individual, or makes choices that only make sense for \emph{this} asker rather than the general public.
            \item \textbf{0} --- Generic: The answer reads as a stock response that could be addressed to almost any user asking the same question. It treats the asker as anonymous --- no reference to their specifics, no individualized framing, no user-conditional reasoning.
        \end{itemize}

        \medskip
        \textbf{Answer:}
    \end{tcolorbox}
    \captionof{figure}{Judge prompt for the \emph{Generic Distinctness} sub-dimension (Persona adaptation).}
    \label{fig:judge_generic_distinctness}
    \vspace{5mm}

    \begin{tcolorbox}[
        colback=gray!10!white, colframe=black, colbacktitle=gray!50!white,
        coltitle=black, rounded corners, boxrule=0.6mm, width=0.92\textwidth,
        fonttitle=\normalsize, fontupper=\footnotesize,
        title={\textbf{Substantive Integration Judge Prompt (Persona Adaptation)}},
    ]
        \textit{You are an expert evaluator for a personalized conversational AI system. Your task is to assess one specific aspect of persona adaptation at a time. Focus only on the requested aspect.}

        \medskip
        \textbf{\# Task Description}\\
        A memory-augmented dialogue system has generated an answer for a user with a hidden \textbf{implicit persona factor}. \\
        Your job: judge \textbf{whether the persona is integrated into the answer's substance} (recommendations, alternatives, caveats, framing, stance) --- beyond a mere surface-level word borrow.

        \medskip
        \textbf{\# Implicit Persona Factor}\\
        \texttt{\{reason\}}

        \medskip
        \textbf{\# Supporting Evidence (relevant past conversations)}\\
        \texttt{\{reference\_conversation\}}

        \medskip
        \textbf{\# Reference Answer}\\
        \texttt{\{gt\_answer\}}

        \medskip
        \textbf{\# Evaluation Target}\\
        Question:~\texttt{\{query\}}\\
        Generated Answer:~\texttt{\{generated\_answer\}}

        \medskip
        \textbf{\# Scoring Rubric --- Substantive Integration (0 / 1)}
        \begin{itemize}[nosep,leftmargin=*]
            \item \textbf{1} --- Integrated: The implicit persona factor shapes at least one substantive content choice in the answer, such as what is recommended, what alternative is offered, what caveat is raised, how the issue is framed, or what stance is taken.
            \item \textbf{0} --- Surface-only or absent: The factor is ignored or only mentioned at the wording/tone level. For example, the answer may say ``given your situation'' or repeat persona-related words, but still give the same recommendation, alternatives, caveats, framing, or stance it would give to any user.
        \end{itemize}

        \medskip
        \textbf{Answer:}
    \end{tcolorbox}
    \captionof{figure}{Judge prompt for the \emph{Substantive Integration} sub-dimension (Persona adaptation).}
    \label{fig:judge_substantive_integration}
\end{figure*}

\begin{figure*}[p]
    \centering
    \begin{tcolorbox}[
        colback=black!10!white, colframe=black, colbacktitle=black!50!white,
        coltitle=black, rounded corners, boxrule=0.6mm, width=0.92\textwidth,
        fonttitle=\normalsize, fontupper=\footnotesize,
        title={\textbf{Shared Prompt Components}},
    ]
        \textbf{\# Node Type Rules}\\
        \textbf{State}: A user condition true at the time it is expressed but subject to change, such as a current stance, goal, active constraint, present situation, or preference.\\
        \textbf{Trait}: A generalized user characteristic that tends to persist across situations and time, such as a stable preference, value, or habitual tendency.\\
        \textbf{Episode}: A summary of what happened in a past conversation, including concrete events, topics, and actions at a particular time.

        \medskip
        \textbf{\# Metadata Rules}\\
        \textbf{scope}: \textsc{BROAD} if the node may affect decisions across multiple future tasks or topics; \textsc{NARROW} if it is mainly tied to the current task, topic, or short-term situation.\\
        \textbf{recall\_priority}: For State nodes, \textsc{HIGH} if ignoring the state would make a response clearly wrong, unsafe, inconsistent with an explicit constraint, or noticeably frustrating; otherwise \textsc{LOW}.

        \medskip
        \textbf{\# Label Rules}\\
        \textbf{keywords}: Specific one-word terms from the source text or close variants.\\
        \textbf{domain\_label}: Broader topical categories that group the node with related nodes.\\
        Each item must be one continuous word with no whitespace. The same string cannot appear in both fields.

        \medskip
        \textbf{\# Evidence Relation Rules}\\
        \textbf{SUPPORT}: The two pieces of information are consistent or mutually reinforcing.\\
        \textbf{CONTRADICT}: The two pieces of information are in clear tension or conflict.\\
        \textbf{SHIFT\_TO}: Older information changes into newer information and is no longer currently valid.\\
        \textbf{IRRELEVANT}: The pair is unrelated or only weakly associated.
    \end{tcolorbox}

    \caption{Shared prompt components reused by extraction, relation classification, and QA generation prompts.}
    \label{fig:shared_prompt_components}
\end{figure*}

\begin{figure*}[p]
    \centering
    \begin{tcolorbox}[
        colback=blue!5!white, colframe=black, colbacktitle=blue!20!white,
        coltitle=black, rounded corners, boxrule=0.6mm, width=0.92\textwidth,
        fonttitle=\normalsize, fontupper=\footnotesize,
        title={\textbf{State Extraction Prompt}},
    ]
        \textit{You are a persona state extraction assistant. Extract user states from the current user turn. Respond in strict JSON.}

        \medskip
        \texttt{\{Node Type Rules\}}\\

        \medskip
        \textbf{\# Task Description}\\
        Extract up to \texttt{\{STATE\_MAX\_COUNT\}} persona \textbf{State} node(s) from the \textbf{CURRENT USER UTTERANCE} only.
        If none, return an empty states list.
        Extract useful persona-relevant signals as States when they are not stable enough to be Traits.
        Do not invent or speculate.

        \medskip
        \textbf{\# Context Usage}\\
        Use prior context and the assistant response only to resolve references.
        Do not extract information stated only outside the current user utterance.

        \medskip
        \textbf{\# Output Format}\\
        Each State must begin with ``The user'', be one concise sentence, and express a currently valid condition, constraint, goal, stance, or preference.
        Rewrite episodic descriptions as implied current States, not raw narration.
        Use placeholder ids \texttt{new\_0}, \texttt{new\_1}, $\ldots$ in extraction order.

        \texttt{\{Metadata Rules\}}\\
        \texttt{\{Label Rules\}}

        \medskip
        \textbf{\# Required Fields}\\
        \texttt{id}, \texttt{content}, \texttt{keywords}, \texttt{domain\_label}, \texttt{scope}, \texttt{recall\_priority}

        \medskip
        \textbf{\# Input}\\
        \texttt{[PRIOR CONTEXT --- disambiguation only; do not extract from here]}\\
        \texttt{\{prior\_context\}}\\
        \texttt{[ASSISTANT RESPONSE --- read-only context; do not extract from here]}\\
        \texttt{\{assistant\_response\}}\\
        \texttt{[CURRENT USER UTTERANCE --- extract from here only]}\\
        \texttt{\{current\_user\_utterance\}}
    \end{tcolorbox}

    \caption{Prompt used to extract time-bounded persona State nodes from the current user utterance.}
    \label{fig:state_extraction_prompt}
\end{figure*}

\begin{figure*}[p]
    \centering
    \begin{tcolorbox}[
        colback=green!5!white, colframe=black, colbacktitle=green!20!white,
        coltitle=black, rounded corners, boxrule=0.6mm, width=0.92\textwidth,
        fonttitle=\normalsize, fontupper=\footnotesize,
        title={\textbf{Episode Extraction Prompt}},
    ]
        \textit{You are an episode extraction assistant. Summarize the recent conversation into one episode. Respond in strict JSON.}

        \medskip
        \texttt{\{Node Type Rules\}}\\

        \medskip
        \textbf{\# Task Description}\\
        Create exactly one \textbf{Episode} from the \textbf{RECENT CONVERSATION}.
        Summarize what happened in the conversation.
        Do not invent or speculate.

        \medskip
        \textbf{\# Extraction Rule}\\
        Do not extract generalized persona traits or separate State nodes.
        Include user traits, preferences, or conditions only when they are part of the concrete episode being summarized.

        \medskip
        \textbf{\# Output Format}\\
        Each Episode must:
        \begin{itemize}[nosep,leftmargin=*]
            \item begin with ``The user'';
            \item be 1--2 sentences;
            \item summarize concrete events, discussed topics, actions, and developments;
            \item describe the episode itself, not generalized persona traits.
        \end{itemize}

        \texttt{\{Metadata Rules\}}\\
        \texttt{\{Label Rules\}}

        \medskip
        \textbf{\# Required Fields}\\
        \texttt{id}, \texttt{content}, \texttt{keywords}, \texttt{domain\_label}, \texttt{scope}

        \medskip
        \textbf{\# Input}\\
        \texttt{[RECENT CONVERSATION]}\\
        \texttt{\{recent\_conversation\}}
    \end{tcolorbox}

    \caption{Prompt used to summarize a recent dialogue chunk into a concrete Episode node.}
    \label{fig:episode_extraction_prompt}
\end{figure*}

\begin{figure*}[p]
    \centering
    \begin{tcolorbox}[
        colback=orange!6!white, colframe=black, colbacktitle=orange!25!white,
        coltitle=black, rounded corners, boxrule=0.6mm, width=0.92\textwidth,
        fonttitle=\normalsize, fontupper=\footnotesize,
        title={\textbf{Trait Extraction Prompt}},
    ]
        \textit{You are a persona trait extraction assistant. Infer at most one new long-term trait from the accumulated recent evidence. Respond in strict JSON.}

        \medskip
        \texttt{\{Node Type Rules\}}\\

        \medskip
        \textbf{\# Task Description}\\
        Infer at most \texttt{\{TRAIT\_MAX\_COUNT\}} persona \textbf{Trait} node(s) from the recent conversations below.
        If the recent conversations do not reveal a clear new persistent pattern, return an empty traits list.
        Do not invent or speculate.

        \medskip
        \textbf{\# Trait Criterion}\\
        Use the ``in general'' test:
        a Trait should still be true if the user is asked about themselves ``in general'' with no specific time, place, or context attached.

        \medskip
        \textbf{\# Output Format}\\
        Each Trait must:
        \begin{itemize}[nosep,leftmargin=*]
            \item begin with ``The user'';
            \item be 2--3 complete sentences;
            \item be inferable as a likely persistent pattern;
            \item not merely describe a one-time event or momentary feeling.
        \end{itemize}

        \medskip
        \texttt{\{Metadata Rules\}}\\
        \texttt{\{Label Rules\}}        

        \medskip
        \textbf{\# Required Fields}\\
        \texttt{id}, \texttt{content}, \texttt{keywords}, \texttt{domain\_label}, \texttt{scope}

        \medskip
        \textbf{\# Input}\\
        \texttt{[Recent conversations]}\\
        \texttt{\{recent\_conversations\}}
    \end{tcolorbox}

    \caption{Prompt used to infer persistent persona Trait nodes from accumulated recent conversations.}
    \label{fig:trait_extraction_prompt}
\end{figure*}

\begin{figure*}[p]
    \centering
    \begin{tcolorbox}[
        colback=red!5!white, colframe=black, colbacktitle=red!20!white,
        coltitle=black, rounded corners, boxrule=0.6mm, width=0.92\textwidth,
        fonttitle=\normalsize, fontupper=\footnotesize,
        title={\textbf{Trait--State Evidence Relation Extraction Prompt}},
    ]
        \textit{You are an evidence classification assistant. Classify direct relationships involving a newly extracted Trait and listed State, Episode, and Trait nodes. Respond in strict JSON.}

        \medskip
        \textbf{\# Shared Rules}\\
        \texttt{\{Node Type Rules\}}\\
        \texttt{\{Evidence Relation Rules\}}

        \medskip
        \textbf{\# Task Description}\\
        Judge direct evidence relationships involving the newly extracted Trait.
        Index 0 is the new Trait.
        Listed State nodes, Episode nodes, and the previous Trait are compared against this new Trait.

        \medskip
        \textbf{\# Pair Types}\\
        Judge each listed State--new Trait pair, each listed Episode--new Trait pair, and the previous Trait--new Trait pair when a previous Trait exists.

        \medskip
        \textbf{\# Direction Rule}
        \begin{itemize}[nosep,leftmargin=*]
            \item For State/Episode $\leftrightarrow$ new Trait pairs: \texttt{source\_id} is the listed State or Episode index, and \texttt{target\_id = 0}.
            \item For previous Trait $\leftrightarrow$ new Trait with \texttt{SHIFT\_TO}: \texttt{source\_id} is the previous Trait index and \texttt{target\_id = 0}, following older information $\rightarrow$ newer replacement.
            \item For cross-type \texttt{SHIFT\_TO}, the source State or Episode provides evidence that updates, replaces, or invalidates the new Trait.
            \item \texttt{SUPPORT}, \texttt{CONTRADICT}, and \texttt{IRRELEVANT} are semantically symmetric, but the specified source--target direction is preserved.
        \end{itemize}

        \medskip
        \textbf{\# Decision Priority}
        \begin{itemize}[nosep,leftmargin=*]
            \item First choose \texttt{IRRELEVANT} if the pair has no clear evidential force.
            \item If related, choose \texttt{SHIFT\_TO} only when the source clearly updates, replaces, or invalidates the target.
            \item If not replacement but clearly in tension, choose \texttt{CONTRADICT}.
            \item If the source grounds, confirms, generalizes into, or reinforces the new Trait, choose \texttt{SUPPORT}.
            \item When uncertain, choose \texttt{IRRELEVANT}.
        \end{itemize}

        \medskip
        \textbf{\# Input}\\
        \texttt{[New Trait]}\\
        \texttt{\{new\_trait\}}\\
        \texttt{[Recent States]}\\
        \texttt{\{recent\_states\}}\\
        \texttt{[Recent Episodes]}\\
        \texttt{\{recent\_episodes\}}\\
        \texttt{[Previous Trait]}\\
        \texttt{\{previous\_trait\}}

        \medskip
        \textbf{\# Output}\\
        Output exactly \texttt{\{N\}} judgments, one for each listed pair in the listed order.
        Use JSON with \texttt{source\_id}, \texttt{target\_id}, and \texttt{relation}.
    \end{tcolorbox}

    \caption{Prompt used to classify evidence relations between a newly extracted Trait and related State, Episode, and previous Trait nodes.}
    \label{fig:trait_state_relation_prompt}
\end{figure*}

\begin{figure*}[p]
    \centering
    \begin{tcolorbox}[
        colback=purple!5!white, colframe=black, colbacktitle=purple!20!white,
        coltitle=black, rounded corners, boxrule=0.6mm, width=0.92\textwidth,
        fonttitle=\normalsize, fontupper=\footnotesize,
        title={\textbf{QA Prompt for Implexconv}},
    ]
        \textit{You are an assistant providing personalized help based on prior conversations with this user.}

        \medskip
        \textbf{\# Shared Rules}\\
        \texttt{\{Node Type Rules\}}

        \medskip
        \textbf{\# Response Rule}\\
        When such a factor applies, reflect it concretely:
        acknowledge the user's goal, name the relevant factor, and adjust the recommendation accordingly.
        Otherwise, answer the question normally without forcing personalization.

        \medskip
        \textbf{\# Conflict Rule}\\
        If the current user message or recent conversation clearly conflicts with, updates, or overrides a retrieved Episode item, the current message takes precedence.

        \medskip
        \textbf{\# Input}\\
        \texttt{[Retrieved States]}\\
        \texttt{\{retrieved\_states\}}\\
        
        \texttt{[Retrieved Traits]}\\
        \texttt{\{retrieved\_traits\}}\\
        
        \texttt{[Retrieved Episodes]}\\
        \texttt{\{retrieved\_episodes\}}\\
        
        \texttt{[Challenged Traits]}\\
        \texttt{\{challenged\_traits\}}\\
        
        \texttt{[Question]}\\
        \texttt{\{question\}}

        \medskip
        \textbf{Answer}: 
    \end{tcolorbox}

    \caption{Prompt used for ImplexConv opposed-subset QA, where retrieved episodes may modify the answer through implicit personalization constraints.}
    \label{fig:qa_opposed_prompt}
\end{figure*}

\end{document}